\documentclass[journal,twoside,web]{ieeecolor}
\usepackage{jsen}
\usepackage{cite}
\usepackage{amsmath,amssymb,amsfonts}
\usepackage{graphicx}
\usepackage{textcomp}
\usepackage{wrapfig}
\usepackage{array}
\usepackage{hhline} 
\usepackage{nomencl}
\usepackage{soul}
\usepackage{xpatch}

\makeatother
\makenomenclature
\renewcommand\nomgroup[1]{\item[\bfseries]}
\def\BibTeX{{\rm B\kern-.05em{\sc i\kern-.025em b}\kern-.08em
    T\kern-.1667em\lower.7ex\hbox{E}\kern-.125emX}}
\newtheorem{definition}{Definition}[section]
\usepackage{ulem}
\usepackage{epstopdf}
\usepackage[justification=centering]{caption}
\usepackage{algorithm}
\usepackage{algorithmicx}
\usepackage{algpseudocode}
\usepackage{array}
\usepackage{float}
\usepackage{subfigure}
\usepackage{cases}
\makeatletter
\newcommand{\Rmnum}[1]{\expandafter\@slowromancap\romannumeral #1@}
\makeatother
\usepackage{makecell}
\newcolumntype{L}[1]{>{\raggedright\let\newline\\\arraybackslash\hspace{0pt}}m{#1}}
\newcolumntype{C}[1]{>{\centering\let\newline\\\arraybackslash\hspace{0pt}}m{#1}}
\newcolumntype{R}[1]{>{\raggedleft\let\newline\\\arraybackslash\hspace{0pt}}m{#1}}
\newtheorem{property}{Property}
\usepackage{cuted}
\usepackage{booktabs}
\usepackage{multirow}
\usepackage{enumerate}
\allowdisplaybreaks[4]
\usepackage{subfloat}
\usepackage{mathrsfs}

\def\BibTeX{{\rm B\kern-.05em{\sc i\kern-.025em b}\kern-.08em
    T\kern-.1667em\lower.7ex\hbox{E}\kern-.125emX}}
\definecolor{abstractbg}{rgb}{0.89804,0.94510,0.83137}
\begin{document}
\title{Learnable WSN Deployment of Evidential Collaborative Sensing Model}
\author{Ruijie Liu, Tianxiang Zhan, Zhen Li, Yong Deng
\thanks{This work was supported in part by the
National Natural Science Foundation of China under Grant 62373078. \textit{(Corresponding author: Yong Deng.)}}
\thanks{Ruijie Liu and Tianxiang Zhan are with the Institute of Fundamental and Frontier Sciences, University of Electronic Science and Technology of China, Chengdu 611731, China (e-mail: liuruijieuestc@hotmail.com, zhantianxianguestc@hotmail.com). }
\thanks{Zhen Li is with China Mobile Information Technology Center, Beijing 100029, China (e-mail: zhen.li@pku.edu.cn).}
\thanks{Yong Deng is with the Institute of Fundamental and Frontier Sciences, University of Electronic Science and Technology of China, Chengdu 611731, China, and also with School of Medicine, Vanderbilt University, Nashville, Tennessee 37240, USA (e-mail: prof.deng@hotmail.com).}}

\IEEEtitleabstractindextext{%
\fcolorbox{abstractbg}{abstractbg}{%
\begin{minipage}{\textwidth}%
\begin{wrapfigure}[12]{r}{3in}%
\includegraphics[width=3in]{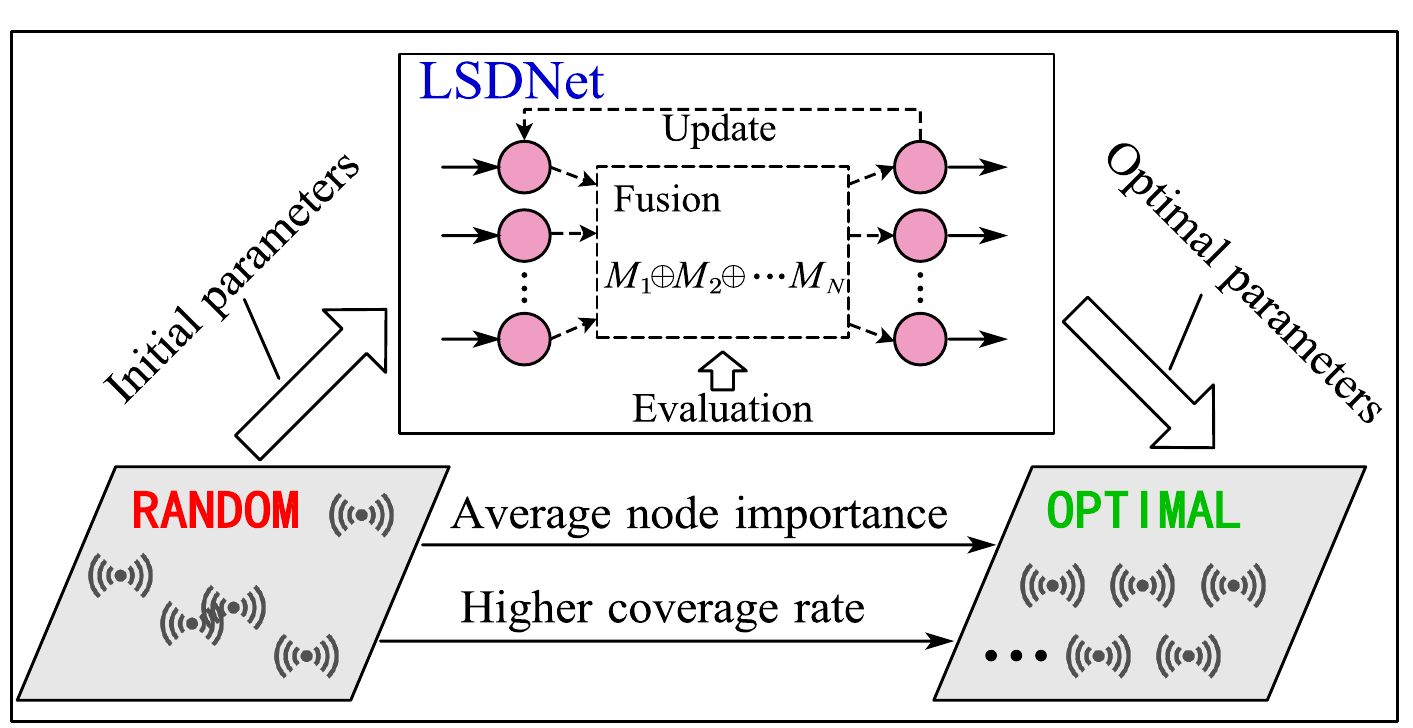}%
\end{wrapfigure}%
\begin{abstract}
In wireless sensor networks (WSNs), coverage and deployment are two most crucial issues when conducting detection tasks. However, the detection information collected from sensors is oftentimes not fully utilized and efficiently integrated. Such sensing model and deployment strategy, thereby, cannot reach the maximum quality of coverage, particularly when the number of sensors within WSNs expands significantly. In this article, we aim at achieving the optimal coverage quality of WSN deployment. We develop a collaborative sensing model of sensors to enhance detection capabilities of WSNs, by leveraging the collaborative information derived from the combination rule under the framework of evidence theory. In this model, the performance evaluation of evidential fusion systems is adopted as the criterion of the sensor selection. A learnable sensor deployment network (LSDNet) considering both sensor contribution and detection capability, is proposed for achieving the optimal deployment of WSNs. Moreover, we deeply investigate the algorithm for finding the requisite minimum number of sensors that realizes the full coverage of WSNs. A series of numerical examples, along with an application of forest area monitoring, are employed to demonstrate the effectiveness and the robustness of the proposed algorithms.
\end{abstract}

\begin{IEEEkeywords}
Wireless sensor networks (WSNs), Deployment, Collaborative sensing model, Evidence theory, Learnable sensor deployment network (LSDNet).
\end{IEEEkeywords}
\end{minipage}}}

\maketitle

\section{Introduction}
\label{sec1}
Wireless sensor networks (WSNs) have attracted thousands of researches over the past decades. The dominant task for a WSN is to carry out the detection of a region of interest, thereby providing the surveillance or perception in environments that are harsh or inaccessible for human beings. For instance, underwater wireless sensor networks are placed in the depths of the sea for the submarine temperature detection, pressure testing, and biological conservation \cite{s18103414}. In the military, enormous tiny sensors are oftentimes deployed to detect the enemy intrusion with the requirements of military operations \cite{SINGH2023118588}. Moreover, WSNs have been extensively integrated into intelligent systems, such as robotics \cite{8334565} and unmanned aerial vehicles (UAVs) \cite{8119562}, \cite{li2025exploring}.

Coverage, deployment, and localization of sensors are three most considerable challenges in WSN applications. Before any task-oriented operation is conducted by a WSN, the self-localization of the unknown nodes being reliant on the anchor nodes, which are equipped with positioning devices such as global positioning system (GPS), is a fundamental and crucial technique \cite{WU20123581}. There are mainly two categories of commonly implemented WSN localization methodologies, namely the range-based localization and the range-free localization. The former enables localization through distance measurements, such as time of arrival (TOA) \cite{8768049}, angle of arrival (AOA) \cite{9785415}, and received signal strength (RSS) \cite{9027850}. The accuracy of the range-based localization is, nevertheless, contingent on the accuracy of the hardware, and on most scenarios, such requirement of high-accuracy results in high energy consumption \cite{chowdhury2016advances}. Instead, a range-free localization method achieves approximate localization outcomes while maintaining low energy consumption by leveraging connectivity information. Therefore, the range-free localization algorithms are widely researched \cite{WU20123581}, \cite{6488858}, particularly in their integration with machine learning methodologies \cite{EZZATIKHATAB2021107915}, \cite{yadav2023optimized}.

Coverage is the most significant issue of a WSN that guarantees the effective detection of a region, which can be normally classified into three types: the area coverage, the target coverage, and the barrier coverage. This paper concentrates on the area coverage, while the target coverage can be deemed as a special case of the area coverage via the transformation using discretized methods. The aim of optimizing the effect of coverage is to enhance criteria like the coverage rate, the energy consumption, and the reliability of coverage to achieve the optimal detection of WSNs. For instance, Banoth et al. \cite{Banoth2023} maximizes the coverage lifetime by sequentially scheduling different sets of sensors. Essentially, such coverage effect is numerically modeled and evaluated by the sensing models of sensors.

In the early research, the Boolean/binary model was widely implemented in WSNs' sensing modeling for its simplicity \cite{1369347}. However, such ideal model fails to properly describe the realistic circumstance in which the detection information is interfered by inevitable uncertain factors, such as noise signals and inherent hardware errors. Therefore, the uncertain sensing model with truncated attenuated detection probabilities was extensively used \cite{DING2023119319}. Moreover, within the framework of transferable belief model (TBM), an evidential sensing model was introduced to quantify detection uncertainties by employing belief functions \cite{SENOUCI2019102414}. From the perspective of estimation and data fusion, Wang et al. \cite{1524581} proposed the concept of the information coverage of WSNs. Furthermore, Deng et al. \cite{6782385} introduced the confident information coverage utilizing the field reconstruction methodology to estimate multi-modal physical attributes. The aforementioned models, nevertheless, partially discard some detection information that can be readily collected, thereby potentially limiting the coverage effect since such information could have been efficiently integrated. To leverage the fusion information of multiple sensors, we consider developing a collaborative sensing model of WSNs under the framework of Dempster-Shafer (D-S) evidence theory, which excels in uncertainty representation and information integration \cite{huang2023higherR}, \cite{zhang2024divergence}, \cite{WANG2024110474}. It is noteworthy that, we design a performance evaluation model of the information fusion system, for rational selection of the sensors to be effectively fused.

Deployment of WSNs is not trivial as vital network performance metrics, including coverage quality, network connectivity, and network lifetime, are critically dependent on it \cite{THIRUMAL2024101165}. If sensors in a WSN are precisely deployed to their locations known in advance, the deployment is called deterministic, which provides the mathematical guarantee for achieving the optimal solution of a WSN deployment problem \cite{dhillon2003sensor}. On most occasions, however, the transportation of sensors to the predetermined locations is unfeasible, thereby the sensors are randomly scattered in the region of detection. In such scenarios, the initial positions of sensors are generated based on the assumption of randomness \cite{SAH202148}. In pursuit of the desirable coverage goal, sensors are relocated to their intended positions through embedded mobile devices \cite{7052338}. In the literature, the problem of the random deployment of WSNs has been demonstrated to be NP-hard \cite{1146711}. Traditional algorithms, thereby, can hardly deal with such problems with limited computational time and resources. Metaheuristic algorithms, such as genetic algorithm (GA) \cite{Njoya2020} and particle swarm optimization (PSO) \cite{8993707}, as alternatives, can be implemented for reaching a suboptimal solution of WSN deployment. Furthermore, Wang et al. \cite{Wang2019} designed a PSO-based optimization scheme of mobile sensors, attempting to address the issue of hole patching of WSNs via relocation. Additionally, enormous studies employing the swarm intelligence algorithms were conducted to achieve both the larger coverage area and the less energy consumption of WSN deployment \cite{ZHANG2023110827}, \cite{10106225}, \cite{cmc201804132}. 

Nevertheless, note that the solution of aforementioned algorithms cannot maintain stable because metaheuristic algorithms inherently own significant stochastic characteristics. Furthermore, the unaffordable computational cost and the frequent convergence to local optimum encountered when addressing large-scale deployment problems, render these algorithms impractical. Thus, the construction of a WSN deployment framework with robustness, coupled with the reduction of the computational cost when solving algorithms, still remains an unresolved issue.

This article aims to establish a learnable sensor deployment network (LSDNet) that resolves the problem of the WSN deployment. The proposed network is rooted in the idea of back propagation of sensors' coordinates, with the intention of equilibrating the importance of sensors within a WSN and achieving the maximum coverage quality. In this network, an evidential collaborative sensing model that utilizes the fusion information of multiple sensors, is proposed to enhance the detection capabilities of WSNs. Moreover, the LSDNet addresses the problem of the substantial time complexity associated with conducting evidential calculations on power sets. The central contributions of this article are previewed as follows.

\begin{enumerate}
\renewcommand{\labelenumi}{\theenumi)}
  \item We develop a LSDNet optimization framework considering both sensor contribution and detection capability, for calibrating sensors' locations via gradient descent.
  
  \item An evidential collaborative sensing model is proposed to enhance the detection capability of WSNs by leveraging the fusion information. 
  
  \item We propose a LSDNet-based algorithm for achieving the optimal WSN deployment, especially in large-scale WSN scenarios. 
  
  \item We tailor a greedy-based algorithm to find the minimum number of sensors that exactly achieves the full coverage of WSNs.
\end{enumerate}

The remaining sections of the article are structured as follows. Section \ref{sec2} symbolically describes the main problem to be resolved, as well as the related formulation in this paper. An evidential collaborative sensing model is put forth in Section \ref{sec3}. Section \ref{sec4} proposes two algorithms based on the LSDNet, aiming to achieve the optimal deployment, as well as find the minimum number of sensors to realize the full coverage. Section \ref{sec5} employs numerical examples and a real-world case to demonstrate the effectiveness of the proposed sensing model and deployment algorithms. Section \ref{sec6} is the conclusion of the whole article.

\section{Problem description and formulation}\label{sec2}
Assume that a set of wireless sensors is deployed in a two-dimensional region $R\in \mathbb{R}^2$, denoted as $S_R = \{s_1, s_2, \ldots, s_K\}$, where $K$ represents the number of sensors.  Targets of detection scattered within the region $R$ are denoted as $T_R = \{t_1, t_2, \ldots, t_N\}$, where $N$ represents the number of targets. The Euclidean distance between the sensor $s_k$ and the target $t_j$ within $R$ is calculated as:
\begin{equation}\label{eq1}
    d(s_k,t_j) = \sqrt{(x_k-x_j)^2+(y_k-y_j)^2}
\end{equation}
where $(x_k, y_k)$ and $(x_j, y_j)$ represent locations of $s_k$ and $t_j$ $(k=1,2,\ldots,K,\ j=1,2,\ldots,N)$, respectively. In this paper, we give the following hypotheses:
\begin{enumerate}
    \item A rectangular partition pattern is employed to divide the continuous detection region into grid points with one-unit interval;
    \item Each sensor is equipped with self-localization devices to obtain the location of the sensor $(x_k, y_k)$;
    \item Each sensor possesses the capability to detect all targets within its sensing range $r_s$ with a probability $p$;
    \item All sensors are homogeneous and movable;   
    \item All computational operations are completed by one or more cluster-heads.
\end{enumerate}

\subsection{Classic Sensing Models}\label{subsec2a}
In early research, Boolean model was frequently employed in the modeling of the sensor perception due to its computational simplicity. 
\begin{definition}[Boolean sensing model]
The probability that the target $t_j$ is detected by the sensor $s_k$ is calculated as \cite{1369347}:
\begin{equation}\label{eq2}
    p\left( s_k,t_j \right) =\left\{ \begin{array}{l}
	1,\quad d\left( s_k,t_j \right) \le r_s\\
	0,\quad d\left( s_k,t_j \right) >r_s\\
\end{array} \right. 
\end{equation}
where $r_s$ is the sensing range of the sensor $s_k$. 
\end{definition}
Nevertheless, the implementation of Boolean sensing model requires ideally interference-free circumstances.

\begin{definition}[Probabilistic sensing model]
To handle the uncertainty in the realistic sensor perception, the probability of detection of sensors is given by \cite{DING2023119319}:
\begin{equation}\label{eq3}
p\left( s_k,t_j \right) =\left\{ \begin{array}{l}
	1,\quad \quad \quad \quad \quad d\left( s_k,t_j \right) <r_s-r_e\\
	e^{(-\frac{\alpha_1\lambda _1^{\beta _1}}{\lambda _2^{\beta _2}}+\alpha _2)},\ r_s-r_e\leq d\left( s_k,t_j \right) <r_s+r_e\\
	0,\quad \quad \quad \quad \quad d\left( s_k,t_j \right) \geq r_s+r_e\\
\end{array} \right. 
\end{equation}
where $\lambda_1=r_e-(r_s-d(s_k,t_j))$ and $\lambda_2=r_e+(r_s-d(s_k,t_j))$. $r_e$ is the uncertain sensing range of the sensor $s_k$. The parameters $\alpha_1$, $\alpha_2$, $\beta_1$, and $\beta_2$ are device-oriented. 
\end{definition}
It bears noting that such model discards certain information when $d(s_k,t_j)\geq r_s+r_e$, while the results of detection may diverge significantly if such information is rationally fused. The evidential collaborative sensing model is, thereby, proposed to enhance the detection capabilities of WSN.

\subsection{D-S Evidence Theory}\label{subsec2b}
D-S evidence theory \cite{dempster2008upper,shafer1976mathematical} is acknowledged as an extension of probability theory, demonstrating superiority in the representation and processing of uncertain information \cite{zhan2024generalized, zhao2024linearity}. This characteristic contributes to its widespread applications in domains of pattern recognition \cite{Xiao2023Acomplexweighted}, \cite{deng2022RPS}, \cite{deng2024RPSR}, emergency management \cite{fei2024novel}, \cite{fei2024dempster}, reliability assessment \cite{https://doi.org/10.1002/qre.3319, chen2024risk}, and quantum computation \cite{Xiao2022GQET, he2024quantumrule}.

\begin{definition}[Frame of Discernment]
Consider an exhaustive event set $X$, which contains $n$ exclusive  elements, is represented as \cite{dempster2008upper}:
\begin{equation}\label{eq4}
	X = \{\xi_1,\xi_2,\ldots,\xi_n\}
\end{equation}
where $X$ is called the frame of discernment (FoD). The corresponding power set $2^{X}$ is represented as \cite{dempster2008upper}:
\begin{equation}\label{eq5}
	\begin{aligned}
	2^{X} = \{\emptyset, \{\xi_1\},\ldots,\{\xi_n\},\{\xi_1, \xi_2\},\ldots,\{\xi_1,\xi_2,...,\xi_n\}\}
	\end{aligned}
\end{equation}
where $\emptyset$ is the empty set.
\end{definition}

\begin{definition}[Mass function]
The mass function $m: \mathbb{R}^{2^X}\rightarrow \mathbb{R}$ denotes the belief representation of the power set of FoD, which satisfies \cite{dempster2008upper}:
\begin{equation}\label{eq6}
    m(F)\in [0,1]
\end{equation}
\begin{equation}\label{eq7}
	\sum_{F \subseteq 2^{X}}{m(F)} = 1
\end{equation}
where $m(F) = 0$ holds if $F=\emptyset$ for a closed framework. Mass function is also called basic probability assignment (BPA), and any subset within $\{F\subseteq 2^X|m(F)>0\}$ is called a focal element.
\end{definition}

\begin{definition}[Dempster combination rule]
For two mass functions $m_1$ and $m_2$ of any element $(F\neq\emptyset)\subseteq 2^X$, the Dempster combination rule is defined as follows \cite{dempster2008upper}:
\begin{equation}\label{eq8}
\begin{aligned}
	m_{1}\oplus m_{2}(F) = \frac{\displaystyle\sum_{A \cap B = F,A,B\subseteq 2^X}{m_1(A)m_2(B)}}{1-Q_{1\oplus2}} 
\end{aligned}
\end{equation}
\begin{equation}\label{eq9}
	Q_{1\oplus2} = m_{1}\oplus m_{2}(\emptyset) = \displaystyle\sum_{A \cap B = \emptyset}{m_1(A)m_2(B)}
\end{equation}
where $Q_{1\oplus2}$ represents the degree of conflict between two information sources.
\end{definition}

\begin{definition}[Uncertainty measures]
Considering an FoD $X$ and the corresponding power set $2^{X}$, the belief entropy and Hartley entropy on this frame are calculated as \cite{Qiang2022fractal}\cite{8116665}:
\begin{equation} \label{eq10}
E_{D}(m) = -\sum\limits_{F \subseteq X} m(F)\log(\frac{m(F)}{2^{|F|}-1})
\end{equation}

\begin{equation} \label{eq11}
E_{H}(m) = \sum\limits_{F \subseteq X} m(F)\log(|F|)
\end{equation}
where $|F|$ represents the cardinality of the subset $F$. The base of the logarithm is taken as 2 in this paper. 
\end{definition}
Comparing with Shannon entropy, the belief entropy and Hartley entropy emphatically consider the non-specificity of multiple subsets, which inherently possess uncertainty within an evidential framework.

\subsection{Performance Evaluation of Evidential Fusion System}\label{subsec2c}
Combination between evidences is deemed as an extension of Baysian inference, which reduces the uncertainty of the original information. To quantitatively evaluate the performance of such evidential fusion, the efficiency of the system is calculated as \cite{LIU2024107262}:
\begin{equation}\label{eq12}
\begin{split}
\eta = 1-\frac{E^{f}_D(m_{1\ldots n})}{E^{g}_D(m_1,\ldots,m_n)}
\end{split}
\end{equation}
where $E^{f}_D(m_{1\ldots n})$ denotes the belief entropy of the fused mass function $m_{1\ldots n}$. $E^{g}_D(m_1,\ldots,m_n)$ denotes the geometric mean belief entropy of the original $n$ masses $m_1,\ldots,m_n$. The range of the efficiency is from $-\infty$ to $1$, where a larger $\eta$ indicates better performance of the evidential fusion.

\section{Evidential Collaborative Sensing Model}
\label{sec3}
In this section, an evidential fusion based collaborative sensing model is put forth to compensate the weak capability of detection of WSNs in uncertain circumstances.
\subsection{Evidential Sensing Model}\label{subsec3a}
\begin{figure}[htpb]
	\centering  
	\includegraphics[width=7cm]{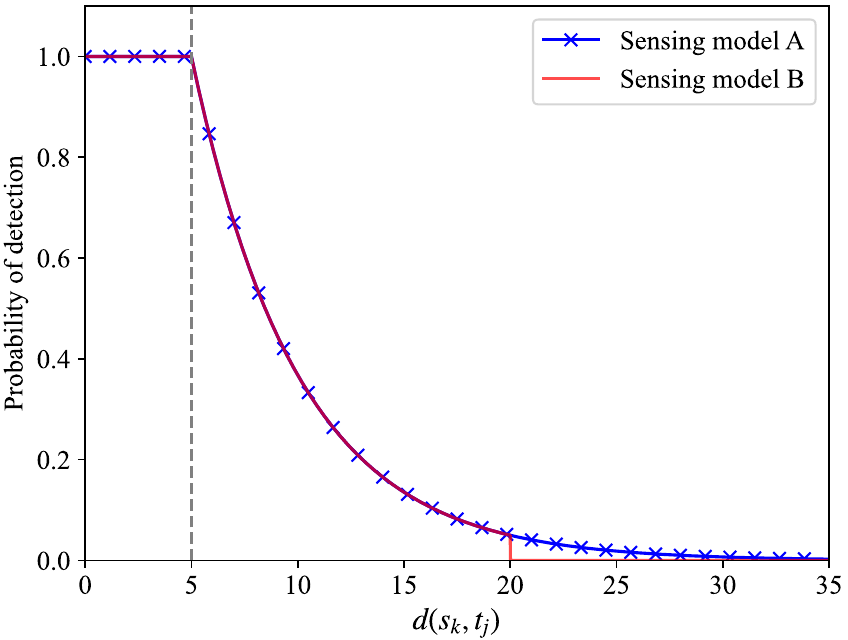}  
	\caption{The probability of detection using two sensing models.}  
	\label{evi_compare}  
\end{figure}

Considering a region $R$ to be detected by a WSN, the probability that the target $t_j$ is detected by the sensor $s_k$ $(j=1,2,\ldots,N,\ k=1,2,\ldots,K)$ within $R$ is computed as:
\begin{equation}\label{eq13}
p_e\left( s_k,t_j \right) =\left\{ \begin{array}{l}
	1,\quad \quad \quad \quad \quad \quad \ \  d\left( s_k,t_j \right) <r_s\\
	e^{-\lambda(d\left( s_k,t_j \right)-r_s)^{\beta}},\ d\left( s_k,t_j \right) \geq r_s\\
\end{array} \right. 
\end{equation}
where $\lambda$ is the coefficient of attenuation. Fig. \ref{evi_compare} depicts the probability of detection employing two sensor sensing models, namely, the proposed evidential sensing model in \eqref{eq13}, denoted as $A$, and the truncated evidential sensing model in \cite{SENOUCI2019102414}, denoted as $B$. From Fig. \ref{evi_compare}, it can be observed that the model $A$ preserves all available information without any truncation.

In D-S evidence theory, the FoD describing whether the target $t_j$ is detected by the sensor $s_k$ is constructed as $X=\{D, ND\}$. Therefore, the mass function of the sensor detection can be defined as:
\begin{equation}\label{eq14}
\begin{aligned}
M_{k,j}&=(m^{\emptyset}_{k,j},m^{D}_{k,j},m^{ND}_{k,j},m^{\{D,ND\}}_{k,j})\\
&=(0,p_e(s_k,t_j),0,1-p_e(s_k,t_j))
\end{aligned}
\end{equation}
where $m^{\emptyset}_{k,j}$ equals to $0$ in a closed framework, and will not be specified in subsequent mass functions. $m^{D}_{k,j}$ and $m^{ND}_{k,j}$ are beliefs that the target $t_j$ is detected and is not detected by the sensor $s_k$, respectively. $m^{\{D,ND\}}_{k,j}$ represents the uncertainty indicating the noises and hardware errors, which is proportional to the sensor-target distance. It is noteworthy that the mass function of detection of a single sensor in \eqref{eq14} will be efficiently integrated using Dempster combination rule and the strategy of the performance evaluation of evidential fusion. Consequently, the detection capability of WSNs can be enhanced by the collaborative sensing model, which will be detailed in Section \ref{subsec3b}.

\subsection{Collaborative Sensing Model Based on Evidential Fusion}\label{subsec3b}
\begin{figure}[htpb]
	\centering  
	\includegraphics[width=6cm]{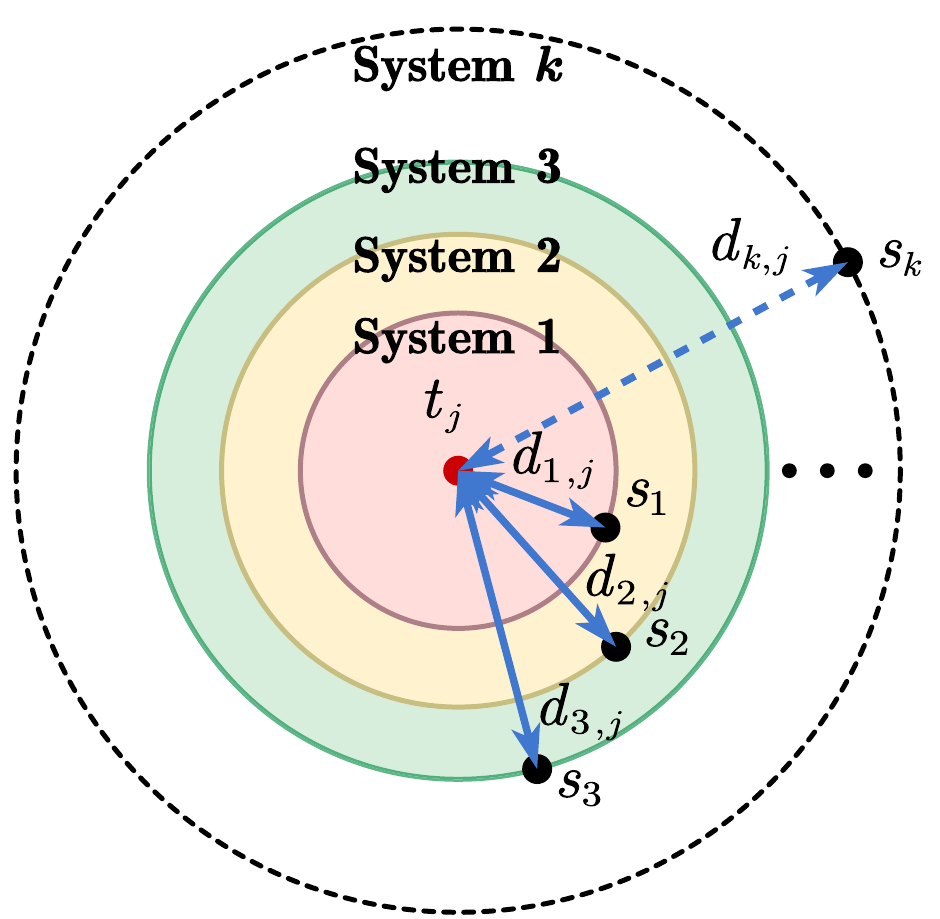}  
	\caption{The collaborative sensing system $k$ to detect the target $t_j$.}  
	\label{sensor_dis}  
\end{figure}

For an evidential fusion system, the fusion efficiency quantifies the ability of uncertainty reduction. Intuitively, the uncertainty reduction of the mass function of the sensor detection replies the enhancement of the detection capability. Assume totally $N$ targets are detected by $K$ sensors in the region of interest $R$. The Euclidean distances between the target $t_j$ $(j=1,2,\ldots,N)$ and sensors are ascendingly vectored as $d_j=\left(d(s_1,t_j),d(s_2,t_j),\ldots,d(s_K,t_j)\right)$. Correspondingly, the mass functions of sensors to detect the target $t_j$ are ascendingly vectored as $M_j=(M_{1,j},M_{2,j},\ldots,M_{K,j})$. Using Dempster combination rule, thereby, the mass function of the first $k$ $(k=1,2,\ldots,K)$ sensors to detect the target $t_j$ is computed as:
\begin{equation}\label{eq15}
\begin{split}
M_{1\ldots k,j} = M_{1,j}\oplus M_{2,j}\oplus\cdots \oplus M_{k,j}
\end{split}
\end{equation}
It is emphasized that  \eqref{eq15} is essentially to fuse the detection results of the first $k$ sensors closest to the target $t_j$. Specifically, the first $k$ sensors constitute the collaborative sensing system $k$ to detect the target $t_j$, which is graphically depicted in Fig. \ref{sensor_dis}. 

To properly evaluate the efficiency of a collaborative sensing system, we construct a two-element framework $\Omega=\{A,B\}$ with $m(A)=1-a$ and $m(\{AB\})=a$, where $0\leq a\leq 1$. The change of the belief entropy and Hartley entropy of the FoD $\Omega$ is illustrated in Fig. \ref{Deng_Hartley}. It can be readily observed that the maximum belief entropy is achieved in point $P$ and the maximum Hartley entropy is achieved in point $Q$. In the proposed evidential sensing model, apparently, the maximum uncertainty of detection is achieved when the target is at infinity from the sensor. Hartley entropy is, thereby, more rational to be implemented for evaluating the performance of the fusion in the evidential sensing model. To detect the target $t_j$, the fusion efficiency of the collaborative sensing system $k$ is computed as:
\begin{equation}\label{eq16}
\eta_H^{k,j} =
	1 - \frac{E_{H}^{f}(M_{1\ldots k,j})}{E_{H}^{g}(M_{1,j},\ldots,M_{k,j})}, \ k=2,3,\ldots,K
\end{equation}
where $E_{H}^{f}(M_{1\ldots k,j})$ is Hartley entropy of the fusion result of mass functions of the first $k$ sensors to detect the target $t_j$ via  \eqref{eq15}. $E_{H}^{g}(M_{1,j},\ldots,M_{k,j})$ is the geometric mean between Hartley entropy of mass functions of the first $k-1$ sensors, and Hartley entropy of the mass function of the $k$th sensor, which is mathematically denoted as:
\begin{figure}
    \centering
    \includegraphics[width=7cm]{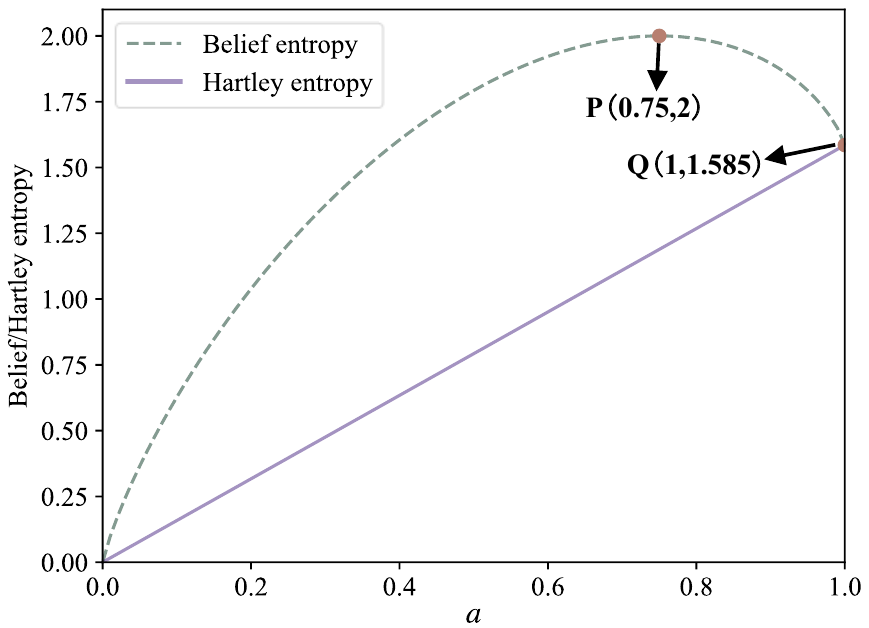}
    \caption{Change of the belief entropy and Hartley entropy with respect to $a$.}
    \label{Deng_Hartley}
\end{figure}
\begin{equation}\label{eq17}
\begin{split}
E_{H}^{g}(M_{1,j},\ldots,M_{k,j}) &= \sqrt{E_{H}^{f}(M_{1\ldots (k-1),j}) \times E_{H}(M_{k,j})}\\
E_{H}^{f}(M_{1,j}) &= E_{H}(M_{1,j})
\end{split}
\end{equation}
where $E_{H}(M_{1,j})$ is Hartley entropy of the mass function of the first sensor, i.e., the one closest to the target $t_j$. Specifically, when $k=1$ and $p_e(s_k,t_j)=1$, we have $\forall k\neq 1$, $\eta_H^{k,j}=0$, because the target $t_j$ is within the sensing range of the sensor $s_k$, i.e., the target $t_j$ is completely detected by a single sensor. In other word, other sensors are redundant in detecting the target $t_j$.

According to the aforementioned context, the collaborative sensing system has the following properties.
\begin{property}{\textbf{\textup{(Uncertainty monotonicity of the mass function of a single sensor).}}}\label{pro1}
The uncertainty of the mass function of a single sensor is nondecreasing when the distance between the sensor and the target increases.
\end{property}
\begin{property}{\textbf{\textup{(Uncertainty monotonicity of the mass function of the collaborative sensing system).}}}\label{pro2}
The uncertainty of the mass function of the collaborative sensing system is nonincreasing when fusing the mass function of any sensor.
\end{property}
\begin{property}{\textbf{\textup{(Boundness of efficiency).}}}\label{pro3}
The range of the efficiency of the collaborative sensing system is $0 \leq \eta^{k,j}_H< 1$. $\eta^{k,j}_H=0$ $(k=2,3,\ldots,K)$ holds if and only if the target $t_j$ is completely detected by the first sensor, i.e., $p_e(s_k,t_j)=1$ for $k=1$. Meanwhile, other sensors do not contribute to the detection.  When $0< p_e(s_k,t_j)<1$ $(k=2,3,\ldots,K)$, $\eta^{k,j}_H=0$ and $\eta^{k,j}_H=1$ cannot be reached.
\end{property}

Detailed proofs of Properties 1-3 are given in Appendices \ref{appa}-\ref{appc}. According to Property \ref{pro3}, a higher efficiency $\eta^{k,j}_H$ indicates a greater gain of detection capability of the collaborative sensing system by fusing the mass function of the sensor $s_k$. The efficiency can be, thereby, employed as a criterion for determining which sensors' mass functions will participate in fusion, i.e., those effectively detect the targets. Accordingly, the number of sensors that \textbf{effectively detect} the target $t_j$ is computed as: 
\begin{equation}\label{eq18}
N_{\text{effect}}^j=\displaystyle\sum_{k=1}^{K}\prod_{g=0}^{k-2}\mathbb{I}(\eta_H^{k-g,j}-\eta_{th})+1, \ k=2,3,\ldots,K
\end{equation}
where $K$ is the total number of sensors. $\eta_{th}$ represents the threshold whether the detection result of a single sensor is effectively fused. $\mathbb{I}(x)$ is an indicator function such that:
\begin{equation}\label{eq19}
\mathbb{I}(x) =\left\{ \begin{array}{l}
	1,\quad x\geq 0\\
	0,\quad x<0\\
\end{array} \right. 
\end{equation}
From  \eqref{eq18} and \eqref{eq19}, at least one sensor effectively detects the target $t_j$, thereby, the range of $N^j_{\text{effect}}$ is $[1,K]$. The first $N_{\text{effect}}^j$ sensors \textbf{collaboratively detect} the target $t_j$, and the corresponding mass function is calculated as:
\begin{equation}\label{eq20}
\begin{split}
M_{1\ldots N_{\text{effect}}^j,j} = M_{1,j}\oplus M_{2,j}\oplus\cdots \oplus M_{N_{\text{effect}}^j,j}
\end{split}
\end{equation}

From  \eqref{eq14} and \eqref{eq20}, the probability $p_e(s_{1\ldots N_{\text{effect}}^j},t_j)$ that the target $t_j$ is detected by the collaborative sensing system $N_{\text{effect}}^j$ can be derived. Based on this, the target $t_j$ is deemed detected if the detection probability of the collaborative sensing system $p_e(s_{1\ldots N_{\text{effect}}^j},t_j)$ satisfies:
\begin{equation}\label{eq21}
\begin{split}
p_e(s_{1\ldots N_{\text{effect}}^j},t_j)\geq p_{th}
\end{split}
\end{equation}
where $p_{th}$ is the probability threshold of detection. Essentially, the proposed collaborative sensing model selects sensors exhibiting sufficient effect on information fusion when detecting targets. In this way, the detection capability of WSNs is enhanced from the perspective of the system sensing. 

\section{A Learnable Framework for WSN Deployment}\label{sec4}
Due to the power-law scale of the evidential framework, the operations conducted on it often possess high computational complexity \cite{ZHAO2022109075}, which results in challenges to resolve optimization problems using metaheuristic algorithms. In the domain of engineering optimizations, the feedback mechanism is extensively adopted for its superiority in finding optima and generalization. In this section, therefore, a learnable framework based on the evidential collaborative sensing model, i.e., the LSDNet, is proposed for achieving an optimal WSN deployment with affordable computational cost, which is discussed in detail in Section \ref{subsec4a}. Moreover, it is imperative to investigate the algorithm concerning the determination of the minimum number of sensors required in a WSN to achieve full coverage using the LSDNet, which is discussed in detail in Section \ref{subsec4b}.

\subsection{Optimal WSN Deployment of LSDNet}\label{subsec4a}
To achieve the optimal deployment of WSNs using a limited number of sensors, the locations of sensors are updated, following the direction of gradient descent to achieve the maximum quality of coverage. Two optimization objectives necessitate consideration, i.e., the maximum coverage rate and the minimum coverage redundancy. In the first place, the coverage rate of WSN deployment stands out as the most essential indicator, satisfying the majority of requirements \cite{9658259}. However, the WSN deployment with such a simplistic objective oftentimes leads to the redundant coverage and the waste of detection capability, potentially resulting in suboptimal quality of coverage. To realize an optimal deployment, the node importance is, thereby, proposed to quantitatively measure each sensor's contribution to detect the targets. Within a WSN, the node importance of the sensor $s_k$ $(k\in \{1,2,\ldots,K\})$ is calculated as:
\begin{equation}\label{eq22}
\begin{split}
NI_k=\sum_{j\in G_k}\frac{p_e(s_{1\ldots N_{\text{effect}}^j},t_j)}{N_{\text{effect}}^j}
\end{split}
\end{equation}
where $G_k = \{j|\prod_{g=0}^{k_j-2}\mathbb{I}(\eta_H^{k_j-g,j}-\eta_{th})>0, 1\leq j\leq N\}$ is the set including the targets that are effectively detected by the sensor $s_k$, and $k_j \in \{2,3,\ldots,K\}$ represents the index of the sensor $s_k$, which is ascendingly arranged based on the distance to the target $t_j$. $p_e(s_{1\ldots N_{\text{effect}}^j},t_j)$ is the probability that the target $t_j$ is detected by the collaborative sensing system $N_{\text{effect}}^j$.
Through the normalization, the node influence $\textbf{NI}\subseteq (\mathbb{R}_+)^K$ is vectored as follows:
\begin{equation}\label{eq23}
\begin{split}
\textbf{NI}=\left(\frac{NI_1}{\sum_{k=1}^{K} NI_k},\frac{NI_2}{\sum_{k=1}^{K} NI_k},\ldots,\frac{NI_K}{\sum_{k=1}^{K} NI_k}\right)
\end{split}
\end{equation}

To achieve the maximum coverage rate, the probability the targets are collaboratively detected is vectored as:
\begin{equation}\label{eq24}
\begin{split}
\textbf{P}_e=\left(p_e(s_{1\ldots N_{\text{effect}}^1},t_1),p_e(s_{1\ldots N_{\text{effect}}^2},t_2),\ldots,p_e(s_{1\ldots N_{\text{effect}}^N},t_N)\right)
\end{split}
\end{equation}

\begin{figure*}
    \centering
    \includegraphics[width=15cm]{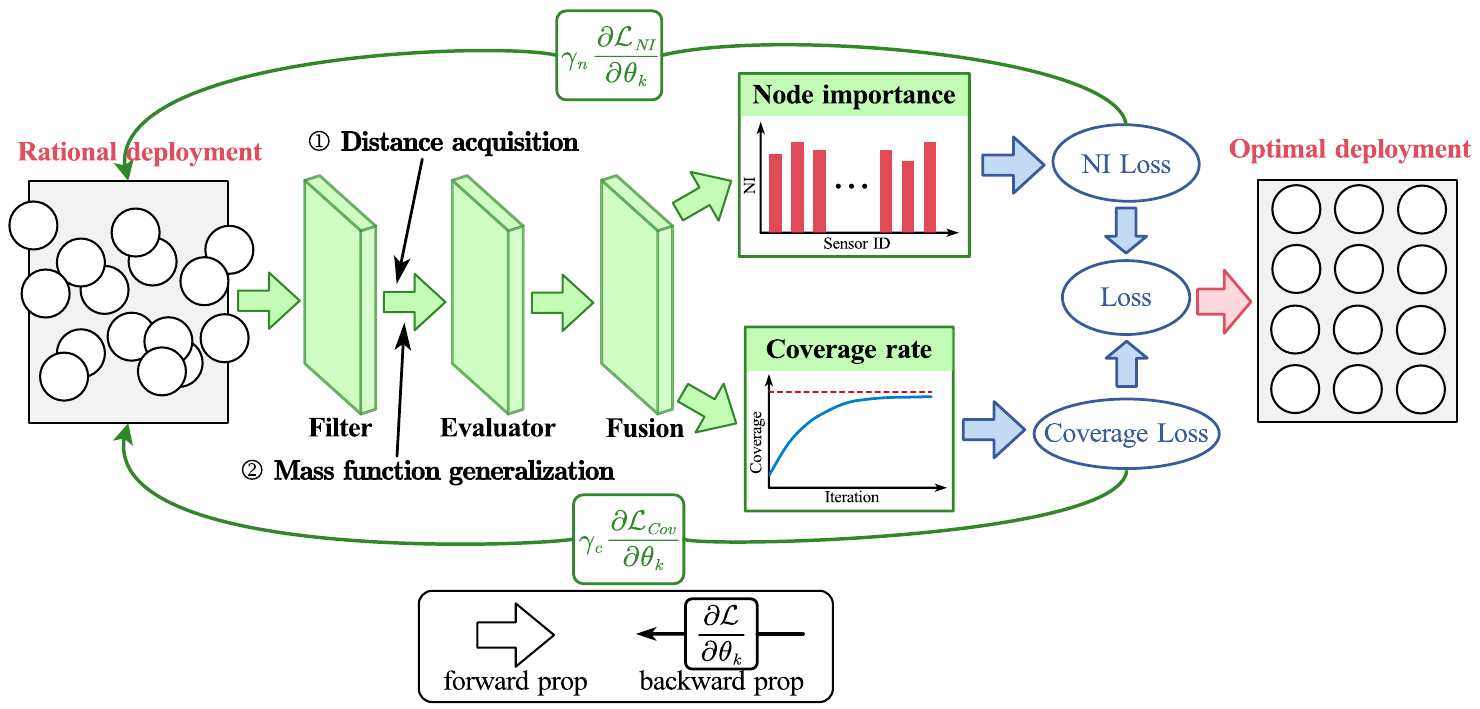}
    \caption{Learnable framework of WSN deployment.}
    \label{framework}
\end{figure*}

It is noteworthy that  \eqref{eq23} and \eqref{eq24} concentrate on evaluations of the coverage quality of WSNs from two perspectives, i.e., the contribution of sensors to detection, and the probability of targets being detected, respectively. For the convenience of calibrating the locations of sensors iteratively, the corresponding coordinates $\Theta_{S_R}\subseteq \mathbb{R}^{2K}$ at $l$th  epoch is denoted as:
\begin{equation}\label{eq25}
\begin{split}
\Theta_{S_R}^{l}&=\left(x_1^{l},y_1^{l},x_2^{l},y_2^{l},\ldots,x_K^{l},y_K^{l}\right)\\
&=\left(\theta_{s_1}^{l},\theta _{s_2}^{l},\theta_{s_3}^{l},\theta_{s_4}^{l},\ldots,\theta_{s_{2K-1}}^{l},\theta_{s_{2K}}^{l}\right)
\end{split}
\end{equation}
where $\theta^l_{s_{2k}}$ and $\theta^l_{s_{2k+1}}$ $(k\in \{1,2,\ldots,K\})$ are horizontal and vertical coordinates of the sensor $s_k$ at $l$th epoch. According to two objectives of optimization, the loss function of the LSDNet to calibrate the sensors' coordinates is defined as:
\begin{equation}\label{eq26}
\begin{split}
\mathcal{L}(\textbf{NI},\textbf{P}_e; \theta_{s_k})=\gamma_{n} \mathcal{L}_{NI}(\textbf{NI}; \theta_{s_k})+\gamma_{c} \mathcal{L}_{Cov}(\textbf{P}_e; \theta_{s_k})
\end{split}
\end{equation}
where $\gamma_{n}$ and $\gamma_{c}$ are hyper-parameters for balancing the loss of node importance and the loss of coverage rate. Such two components of loss are computed via mean square error, namely:
\begin{equation}\label{eq27}
\begin{split}
\mathcal{L}_{NI}(\textbf{NI}; \theta_{s_k})&=MSE(\textbf{NI},\textbf{P}_{\text{uniform}})\\
&=\frac{1}{K} \sum_{k=1}^K (NI_k - \frac{1}{K})^2
\end{split}
\end{equation}
\begin{equation}\label{eq28}
\begin{split}
\mathcal{L}_{Cov}(\textbf{P}_e; \theta_{s_k})&=MSE(\textbf{P}_e,p_{max})\\
&=\frac{1}{N} \sum_{j=1}^N (p_e(s_{1\ldots N_{\text{effect}}^j},t_j) - 1)^2
\end{split}
\end{equation}
where $\textbf{P}_{\text{uniform}}$ represents the uniform distribution, indicating the equivalence of the node influence within a WSN. $p_{max}$ is the maximum probability 1 of detection. The gradient of the loss function at $l$th epoch is computed as:
\begin{equation}\label{eq29}
\begin{split}
\nabla\mathcal{L}^l(\theta_{s_k}) = \gamma_n\frac{\partial \mathcal{L}_{NI}^l(\theta_{s_k})}{\partial \theta_{s_k}^l}+\gamma_c\frac{\partial \mathcal{L}_{Cov}^l(\theta_{s_k})}{\partial \theta_{s_k}^l}
\end{split}
\end{equation}
It is noteworthy that the coordinates of sensors are regarded as the parameters to be updated within the LSDNet, which participate in the forward and backward propagation of the network. On the whole, the forward propagation of the LSDNet encompasses three layers, i.e., the filter, the evaluator, and the fusion layer. Firstly, the filter limits sensors' coordinates to ensure they do not exceed the boundary of the region of interest. Then, the evaluator calculates the performance of collaborative sensing systems, which encompasses following three modules:
\begin{enumerate}
    \item Distance measurement between sensors and targets;
    \item Generalization of sensors' mass function of detection;
    \item Evaluation of the performance of collaborative sensing systems and the identification of the effective nodes.
\end{enumerate}
\color{black}
Lastly, the fusion layer fuses mass functions of sensors that conduct collaborative detection, and obtains the detection results. The overall framework of the LSDNet is depicted in Fig. \ref{framework}. In the backward propagation, the loss function of node importance and the loss of coverage rate are calculated, which characterizes the contribution of sensors to detection and the coverage quality, respectively. Through the forward and backward propagation, the fused information of detection is learned to optimize sensors' locations which minimizes the loss function. The pseudo code of the LSDNet-based optimization of WSN deployment is given as Algorithm \ref{alg1}.
\begin{algorithm}[!t]\small
	\caption{\small{LSDNet-based optimization of WSN deployment.}}
	\label{alg1}
	\begin{algorithmic}[1]
		\Require The initial coordinate of sensors $\Theta_{S_R}^{1}$; The coordinate of targets $\Theta_t$; The maximum epoch $Epoch_{max}$; The threshold of efficiency $\eta_{th}$; The number of sensors and targets $K$ and $N$; The discount coefficients $\gamma_{n}$ and $\gamma_{c}$; The learning rate $\zeta$.
		\Ensure The optimal coordinate of sensors $\Theta_{S_R}^{*}$. 
		\For {$l\gets1$ to $Epoch_{max}$}
        \State Filter the coordinate of sensors;
        \For {$each\ sensor\ s_k \in S_R$}
        \For {$each\ target\ t_j \in T_R$}
		\State Calculate the sensor-target distance $d(s_k,t_j)$ by  \eqref{eq1};
        \State Generate the mass function of detection $M_{k,j}$ by 
        \Statex \qquad \qquad \enspace \eqref{eq13} and \eqref{eq14};
        \State Calculate the efficiency $\eta_H^{k,j}$ by  \eqref{eq15}, \eqref{eq16}, and \eqref{eq17};
        \State Determine the number of sensors $N_{\text{effect}}^j$ effectively
        \Statex \qquad \qquad \enspace detecting the target $t_j$ by  \eqref{eq18} and \eqref{eq19};
        \State Calculate the mass function of the collaborative sens-
        \Statex \qquad \qquad \enspace ing system $M_{1\ldots N_{\text{effect}}^j,j}$ to defect the target $t_j$ by 
        \Statex \qquad \qquad \enspace \eqref{eq20};
        \EndFor
        \State Calculate the loss of node importance and coverage rate
        \Statex \qquad \quad by  \eqref{eq27} and \eqref{eq28};
        \State Update the coordinate of sensors by $\theta_{s_k}^l\gets\theta_{s_{k-1}}^l-$
        \Statex \qquad \quad $\zeta\nabla\mathcal{L}^l(\theta_{s_k})$;
        \EndFor
		\EndFor
	\end{algorithmic}
\end{algorithm}

\subsection{Minimum Sensors Acquisition for WSN Full Coverage}\label{subsec4b}
In WSN applications, the issue of determining the requisite minimum number of sensors to achieve the full coverage of a region of interest, is of significant concern. It has been demonstrated the minimum number of sensors for full coverage in a rectangular region in \cite{10.1145/1978802.1978811}. The full coverage using evidential collaborative sensing model, however, fails to possess fixed minimum number of sensors due to the interactions among multiple sensors. Moreover, it can hardly be found the minimum number of sensors of an irregular region by traditional methods. The LSDNet, in such instances, can be tailored for resolving the aforementioned problems. 

The underlying idea of the proposed algorithm is rooted in the greedy algorithm, which iteratively removes redundant sensors to the greatest extent under the promise of achieving the full coverage using the LSDNet. Mathematically, considering a two-dimensional region $R$ with an irregular boundary to be fully covered by a WSN, the minimum number of sensors by employing the LSDNet is denoted as $K^{min}_R$. The region framed by the minimum bounding rectangle of the region $R$ is denoted as MBR($R$), and the minimum number of sensors for achieving the full coverage of this region is denoted as $K_{MBR}^{min}$. Such number can be determined by the geometric method introduced in \cite{10.1145/1978802.1978811}. For the rectangular partition pattern of sensor deployment, specifically, the optimal interval between sensors is $\sqrt{2}r_s$, and $K_{MBR}^{min}$ is calculated as:
\begin{equation}\label{eq30}
\begin{split}
    K_{MBR}^{min}=\lceil \frac{L_{MBR}}{\sqrt{2}r_s}\rceil \times \lceil\frac{W_{MBR}}{\sqrt{2}r_s}\rceil
\end{split}
\end{equation}
where $\lceil x \rceil$ represents rounding up to the nearest integer for $x$. $L_{MBR}$ and $W_{MBR}$ represent the length and width of the region MBR($R$), respectively. It can be readily proved that $K_{MBR}^{min}$ is a suboptimal solution of the minimum sensors acquisition algorithm for WSN full coverage of the region $R$. In other word, $K_{MBR}^{min}$ is an upper bound of $K^{min}_R$, therefore it can be employed as an initial solution when solving this algorithm.

To measure the redundancy of sensors, it is primary to quantify the overlap degree of sensor coverage in the area to be detected. Specifically, the set of neighbors of the sensor $s_k$ $(k\in \{1,2,\ldots,K\})$ is denoted as:
\begin{equation}\label{eq31}
\begin{split}
S^{k}_{\text{neighbor}}=\{s_{k'}|s_{k'}\in S_R, 0\leq d(s_k,s_{k'})\leq r_a,k'\neq k\}
\end{split}
\end{equation}
where $S_R$ denotes the set containing all sensors. $d(s_k,s_{k'})$ denotes the distance between the sensor $s_k$ and the sensor $s_{k'}$. The overlapping radius $r_a$ must be larger than the sensor interval that exactly achieves the full coverage of the region without collaborative sensing capacities. Specifically, $r_a> \sqrt{2}r_s$ for the rectangular partition pattern of sensor deployment. According to the number of neighbor sensors, the sensor $s_k$ is called \textbf{redundant} if $S^{k}_{\text{neighbor}}\neq \emptyset$, i.e., the cardinality $|S^{k}_{\text{neighbor}}|\neq 0$. For instance, Fig. \ref{overlap} depicted the sensor $s_k$ is surrounded by four neighbor sensors within its overlapping radius. The set of its neighbor sensors is denoted as $S^{k}_{\text{neighbor}}=\{s_1,s_2,s_3,s_4\}$, and $|S^{k}_{\text{neighbor}}|=4$. Therefore, the sensor $s_k$ is deemed redundant and necessitates removal. According to  \eqref{eq25}, the coordinate of sensors removing the sensor $s_k$ is denoted as:
 \begin{figure}[htpb]
 \label{fig5}
	\centering  
	\includegraphics[width=7cm]{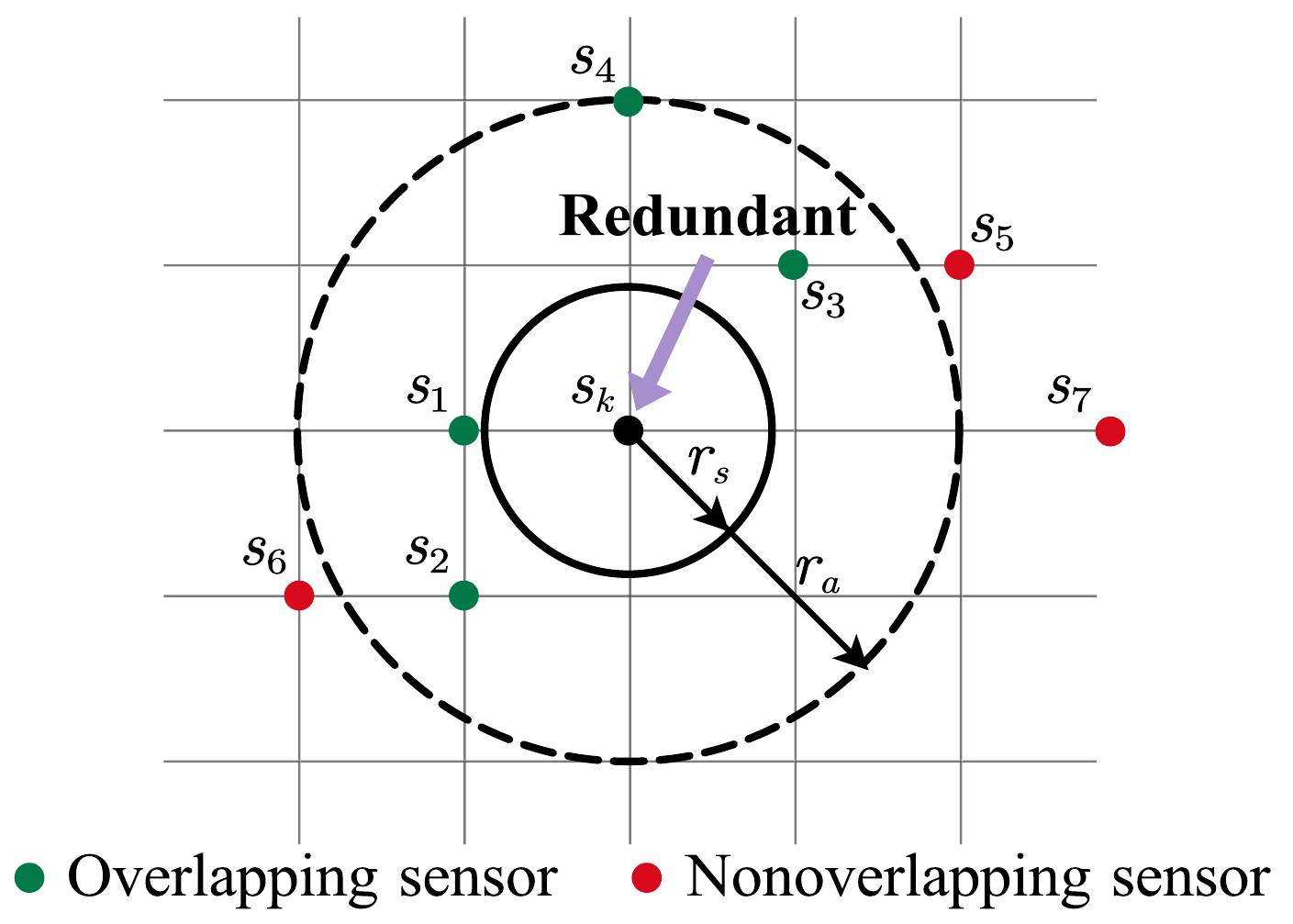}  
	\caption{An example of sensors overlap.}  
	\label{overlap}  
\end{figure}
\begin{equation}\label{eq32}
\begin{split}
\hat{\Theta}_{S_R \backslash s_k}=\left(\theta_{s_1},\theta _{s_2},\ldots,\theta_{s_{2k-2}},\theta_{s_{2k+1}},\ldots,\theta_{s_{2K-1}},\theta_{s_{2K}}\right)
\end{split}
\end{equation}

Utilizing  \eqref{eq32}, the redundant sensors with the most neighbors are iteratively removed until all sensors have no neighbor, thereby achieving full coverage of the region $R$ with the minimum number of sensors. The pseudo code of the minimum sensors acquisition for WSN deployment is given as Algorithm \ref{alg2}.
\begin{algorithm}[!ht]\small
	\caption{\small{Minimum sensors acquisition for WSN deployment using the LSDNet.}}
	\label{alg2}
	\begin{algorithmic}[1]
		\Require The set of initial sensors $S_R$; The coordinate of initial sensors $\Theta_{S_R}$; The overlapping radius $r_a$.
		\Ensure The coordinate of sensors after removing redundant sensors $\hat{\Theta}_{S_R}$.
    \State Calculate the minimum number of sensors $K_{MBR}^{min}$ of the region MBR($R$);
	\Repeat
      \State Use the LSDNet to find the optimal coordinate of sensors
      \Statex \quad \enspace $\Theta_{S_R}^{*}$ by Algorithm \ref{alg1};
     \Repeat
    \State Determine the redundant sensor $s_k$ which satisfies 
    \Statex \qquad \enspace \ $\left\{s_k|\max\limits_{k=1,2,\ldots,K}|S^{k}_{\text{neighbor}}|, s_k\in S_R\right\}$;
    \State Update the coordinate of sensors $\hat{\Theta}_{S_R}\gets\hat{\Theta}_{S_R \backslash s_k}$ by
    \Statex \qquad \quad  \eqref{eq32};
    \State Update the set of sensors $S_R\gets S_R \backslash s_k$;
        \Until{$S^{k}_{\text{neighbor}}=\emptyset$}
    \Until{$|\hat{\Theta}_{S_R}|$=$|\Theta_{S_R}^{*}|$}
	\end{algorithmic}
\end{algorithm}

\section{Applications}
\label{sec5}
In order to demonstrate the superiority of the proposed LSDNet-based algorithm of WSN deployment in the coverage realization and computational cost, a series of comparative numerical examples are conducted. Moreover, a real-world monitoring application is used to illustrate the effectiveness of the minimum sensors acquisition algorithm. It is noteworthy that all experiments are conducted when the sensors are deployed in two-dimensional regions without considering strong signal interference. All experiments are run by Python 3.8.6 on an AMD EPYC 7742 2.25GHz CPU, and an NVIDIA RTX 3080 Ti (12GB), and 40GB RAM.

\subsection{Numerical Examples}\label{subsec5a}
\subsubsection{Comparative Experiments}
Assume a square region $\mathbb{R}_q^2$ with the side length $q$ ranging from 50-200 meters (abbreviated as: m), is to be detected by a WSN comprising dozens of sensors. The entire region is partitioned into a grid of $(q+1) \times (q+1)$ discrete nodes, each separated by the unit length, which uniquely determines the position of sensors. To achieve the optimal deployment of WSNs, the evaluation of deployment algorithms concentrates on two perspectives, namely the coverage rate and the time consumption. Such two indicators correspond to the effectiveness and the practicality of these algorithms in the increasingly large-scale and intricate deployment of WSNs, respectively. Mathematically, the coverage rate of a WSN is computed by:
\begin{equation}\label{eq33}
\begin{split}
\rho = \frac{N_{\text{detected}}}{(q+1)^2}
\end{split}
\end{equation}
where $N_{\text{detected}}$ denotes the number of targets being detected, which satisfies $N_{\text{detected}}\leq (q+1)^2$.

It is worth noting that the experiment encompasses configurations involving the deployment of the maximum of 20, 50, and 100 sensors within square regions with the side length $q$ of 50 m, 100 m, and 200 m, respectively. Table \ref{tab1} gives the detailed parameters of the LSDNet-based optimization algorithm of WSN deployment. Four mainstream classical algorithms, namely, particle swarm optimization (PSO) \cite{SINGH2021100342}, virtual force-directed particle swarm optimization (VFPSO) \cite{s7030354}, individual particle optimization (IPO) \cite{salehizadeh2010coverage}, and glowworm swarm optimization (GSO) \cite{LIAO201112180}, are implemented for comparison in both coverage rate and time consumption. Probabilistic sensing model of sensors is employed in these four algorithms. To illustrate the collaborative effect of the proposed sensing model, the device-dependent parameters of the probabilistic sensing model are set as follows. The parameters $\alpha_1$, $\alpha_2$, $\beta_1$, and $\beta_2$ in  \eqref{eq3} are set to 0.07, 0, 1, and 0, respectively. The sensing range of sensors is set to 8 m. The uncertain sensing range of sensors is set to 4 m. The configuration ensures equivalent detection capabilities for a single sensor between the probabilistic sensing model and the proposed collaborative evidential sensing model. The remaining parameters are consistent with the values outlined in Table \ref{tab1}.
\begin{table}[h]
    \centering
    \caption{Parameters of the LSDNet-based optimization of WSN deployment.}
    \begin{tabular}{lc}
    \toprule[0.8pt]
        Parameters & Values \\
    \midrule
        Deployment pattern & Random\\
        Seed & Random\\
        Sensing range of sensors & $4$ m\\
        Number of sensors & $20-100$\\
        MAC protocol & TDMA\\
        Training epoch & $100-8000$\\
        Optimizer & Adam\\
        $\lambda$ & $0.07$\\
        $\beta$ & $1$\\
        $p_{th}$ & $0.8$\\
        $\eta_{th}$ & $0.2$\\
        $\gamma_n$ & $3\times 10^5$\\
        $\gamma_c$ & $1\times 10^3$\\
        Learning rate & $3\times 10^{-2}$\\
    \bottomrule[0.8pt]
    \end{tabular}
    \label{tab1}
\end{table}

\begin{figure*}[t]
	\centering
    \includegraphics[width=\textwidth]{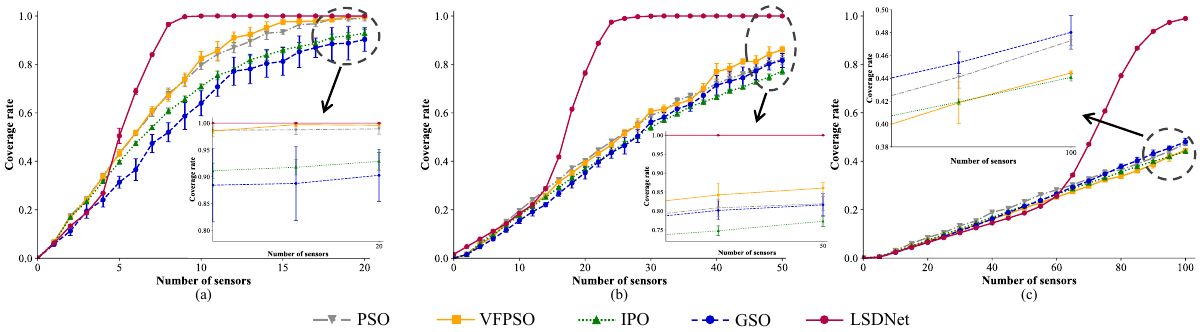}  
	\caption{Coverage rate of WSN deployment with the increase of number of sensors on different areas: (a)$50\times 50 \ \text{m}^2$; (b)$100\times 100 \ \text{m}^2$; (c)$200\times 200 \ \text{m}^2$.}
	\label{fig6}
\end{figure*}

In this section, all experiments were independently executed 10 times. Fig. \ref{fig6} depicts the coverage rate with standard deviation by implementing five
algorithms to deploy a WSN in different areas. As indicated in Fig. \ref{fig6}, when the number of sensors to be deployed increases, the five algorithms of WSN deployment yield higher coverage rates. It is obvious that the LSDNet-based algorithm outperforms PSO, VFPSO, IPO, and GSO, with respect to both coverage rate and the stability of solutions. This is because the collaborative detection information of sensors enormously enhances the detection capabilities of the WSN. Traditional optimization algorithms, such as PSO and VFPSO, also achieve the maximum coverage rate when deploying a small number of sensors in a small-scale region. These algorithms, however, can hardly manage the large-scale issue of WSN deployment. On the one hand, the complex nonlinear mapping between the coordinate of sensors and the effect of coverage, when applying stochastic optimization algorithms, oftentimes results in the instability of solutions. On the other hand, the excessive decision variables yield the intractability for finding the globe optimal solution, accompanied by unaffordable computational cost. Furthermore, Table \ref{tab2} gives the detailed values of coverage rate and time consumption by setting the configurations of:
\begin{itemize}
    \item Case \Rmnum{1}: 20 sensors, $50\times 50 \ \text{m}^2$ area;
    \item Case \Rmnum{2}: 50 sensors, $100\times 100 \ \text{m}^2$ area;
    \item Case \Rmnum{3}: 100 sensors, $200\times 200 \ \text{m}^2$ area.
\end{itemize}
\begin{table*}[ht]
\centering
\caption{Results of WSN deployment using different algorithms.}
\label{tab2}
\renewcommand{\arraystretch}{1}
\begin{tabular}{llcccccc}
\toprule[0.8pt]
\multirow{2}{*}[-0.8ex]{Algorithms} & &\multicolumn{2}{c}{Case \Rmnum{1}} &\multicolumn{2}{c}{Case \Rmnum{2}} &\multicolumn{2}{c}{Case \Rmnum{3}}\\
\cmidrule(r){3-4} \cmidrule(r){5-6} \cmidrule(r){7-8}
& &Coverage rate &Time (unit: s) &Coverage rate &Time (unit: s) &Coverage rate &Time (unit: s)\\
\midrule
\multirow{4}{*}{PSO \cite{SINGH2021100342}} &Max& $99.88\%$&$19.08$&$85.75\%$&$186.82$&$47.96\%$&$2136.81$\\
                     &Ave&$98.94\%$&$18.61$&$81.92\%$&$186.28$&$47.29\%$&$2105.56$\\
                     &Min&$96.66\%$&$18.29$&$79.36\%$&$185.52$&$46.55\%$&$2074.55$\\
                     &SD&$0.99\%$&$0.19$&$1.88\%$&$0.45$&$0.43\%$&$21.98$\\[1ex]

\multirow{4}{*}{VFPSO \cite{s7030354}} &Max& $100.00\%$&$23.89$&$88.29\%$&$323.54$&$44.90\%$&$2632.45$\\
                     &Ave&$99.61\%$&$23.83$&$86.05\%$&$300.57$&$44.48\%$&$2567.97$\\
                     &Min&$98.63\%$&$23.78$&$84.67\%$&$260.79$&$44.27\%$&$2497.57$\\
                     &SD&$0.51\%$&$0.03$&$1.43\%$&$24.80$&$0.18\%$&$44.23$\\[1ex]

\multirow{4}{*}{IPO \cite{salehizadeh2010coverage}} &Max& $96.89\%$&$0.99$&$79.22\%$&$12.29$&$44.81\%$&$98.35$\\
                     &Ave&$92.88\%$&$0.98$&$77.30\%$&$11.86$&$44.09\%$&$95.70$\\
                     &Min&$91.20\%$&$0.97$&$75.42\%$&$11.59$&$43.65\%$&$90.20$\\
                     &SD&$1.62\%$&$0.01$&$1.35\%$&$0.22$&$0.32\%$&$2.51$\\[1ex]
                     
\multirow{4}{*}{GSO \cite{LIAO201112180}} &Max& $96.16\%$&$0.76$&$86.61\%$&$7.70$&$50.96\%$&$72.08$\\
                     &Ave&$90.28\%$&$0.74$&$81.62\%$&$7.39$&$48.02\%$&$68.30$\\
                     &Min&$80.55\%$&$0.69$&$77.20\%$&$7.15$&$46.23\%$&$65.91$\\
                     &SD&$4.86\%$&$0.02$&$2.92\%$&$0.16$&$1.47\%$&$1.66$\\[1ex]

\multirow{4}{*}{LSDNet-based} &Max& $\textbf{100.00\%}$&$6.43$&$\textbf{100.00\%}$&$9.69$&$\textbf{99.28\%}$&$\textbf{32.53}$\\
                     &Ave&$\textbf{100.00\%}$&$5.96$&$\textbf{100.00\%}$&$9.09$&$\textbf{98.96\%}$&$\textbf{32.28}$\\
                     &Min&$\textbf{100.00\%}$&$5.68$&$\textbf{100.00\%}$&$8.13$&$\textbf{98.64\%}$&$\textbf{31.91}$\\
                     &SD&$\textbf{0}$&$0.19$&$\textbf{0}$&$0.51$&$\textbf{0.17\%}$&$\textbf{0.21}$\\
\bottomrule[0.8pt]
\end{tabular}
\end{table*}

In Table \ref{tab2}, the data of each algorithm comprises the maximum (Max), average (Ave), minimum (Min), and standard deviation (SD) values, which are sequentially tabulated from the first to the last row. For Case \Rmnum{1}, IPO and GSO have the lowest cost in time consumption, because they do not involve the optimization of swarm. The time consumption of the LSDNet-based deployment is between the aforementioned two algorithms, and the swarm optimization algorithms. It reveals that meta-heuristic algorithms are relatively practical in small deployment issues. From the perspective of coverage effect, nevertheless, the LSDNet-based deployment achieves the maximum coverage rates in all cases. Moreover, the computational time of the LSDNet-based algorithm exhibits a moderate growth pattern along with the incremental number of sensors and targets of detection, whereas that of swarm optimization algorithms becomes unfeasible. These results demonstrate that the LSDNet-based algorithm of WSN deployment exhibits superiority in improving coverage rate, reducing time consumption, and ensuring solution stability, particularly in large-scale scenarios.

\subsubsection{Sensitivity Analysis}
In real-world WSN deployment, the scatter of WSNs is relatively stochastic because of the interference from uncertain natural factors \cite{059905}. Therefore, the coordinates of sensors oftentimes follow a specific distribution, which can be determined by a probability density function (PDF). In this section, we randomly deploy 100 sensors into a $200\times 200\ \text{m}^2$ square area. To demonstrate the robustness of the LSDNet-based algorithm of WSN deployment, the diverse initial deployment patterns are implemented for the optimization of sensor coordinates, which are tabulated in Table \ref{tab3}. 
\begin{table}[ht]
    \centering
    \caption{Different initial patterns of WSN deployment and their PDFs.}
    \begin{tabular}{lc}
    \toprule[0.8pt]
        Deployment patterns & PDFs \\
    \midrule
        Centroid deployment & / \\
        Boundary deployment & / \\
        Gaussian deployment & $f(x)=\frac{1}{\sqrt{2\pi}\sigma_G}e^{-\frac{(x-\mu)^2}{2\sigma_G^2}}$\\
        Logistic deployment & $f\left( x \right) =\frac{e^{-\left( x-\mu \right) /\sigma_L}}{\sigma_{L}\left( 1+e^{-\left( x-\mu \right) /\sigma_L} \right) ^2}$\\
        Uniform deployment & $f(x)=\frac{1}{b-a}$\\
        Exponential deployment & $f(x)=\varLambda e^{-\varLambda x}$\\
    \bottomrule[0.8pt]
    \end{tabular}
    \label{tab3}
\end{table}

\begin{figure*}[h]
		\centering  
		\subfigtopskip=2pt 
		\subfigbottomskip=6pt 
		\subfigcapskip=-5pt 
		\subfigure[Centroid deployment (initial)]
		{
			\label{centroid_ini}
			\includegraphics[width=0.22\linewidth]{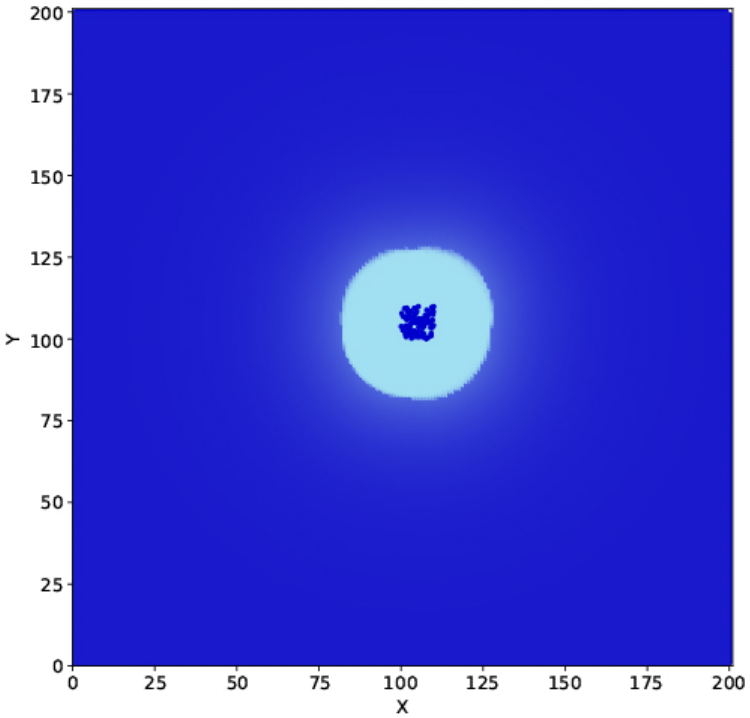}
		}\hspace{-0.005\linewidth}
		\subfigure[Centroid deployment (optimal)]
		{
			\label{centroid_opt}
			\includegraphics[width=0.22\linewidth]{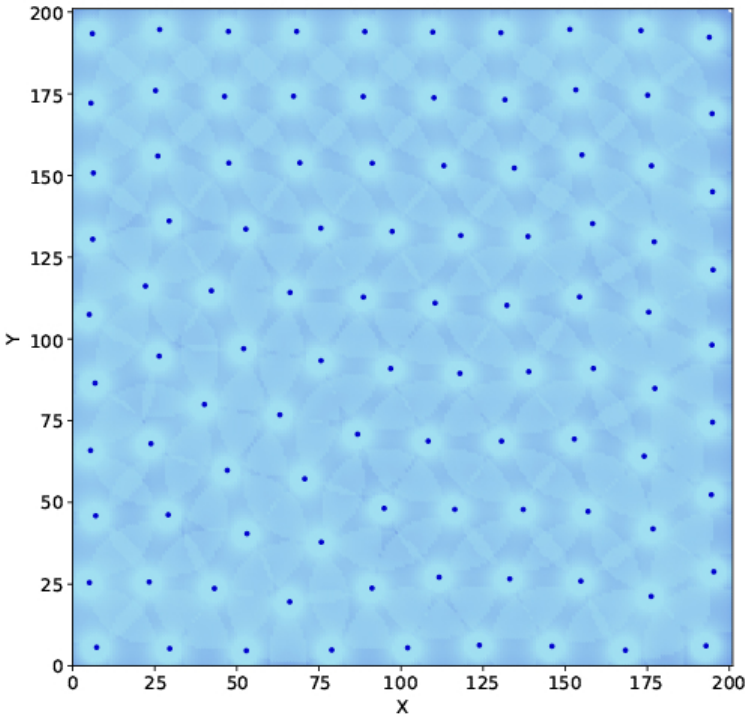}
		}\hspace{-0.005\linewidth}
		\subfigure[Boundary deployment (initial)]
		{
			\label{boundary}
			\includegraphics[width=0.22\linewidth]{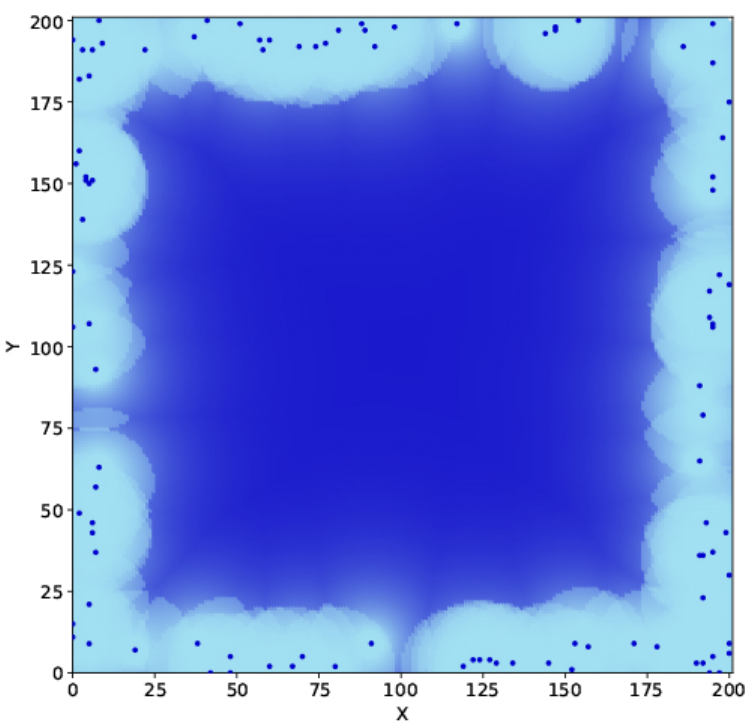}
		}\hspace{-0.005\linewidth}
		\subfigure[Boundary deployment (optimal)]
		{
			\label{boundary_opt}
			\includegraphics[width=0.22\linewidth]{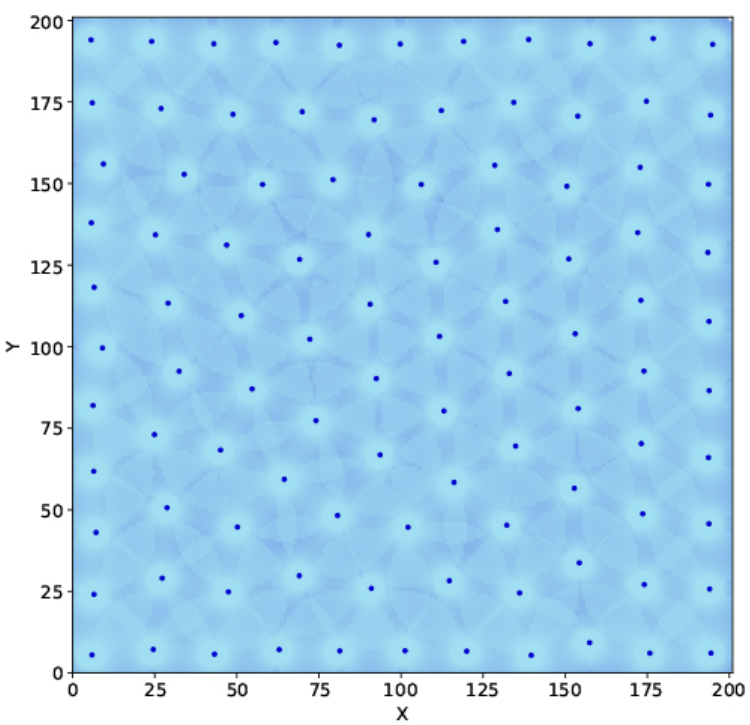}
		}
		\subfigure[Gaussian deployment (initial)]
		{
			\label{gaussian}
			\includegraphics[width=0.22\linewidth]{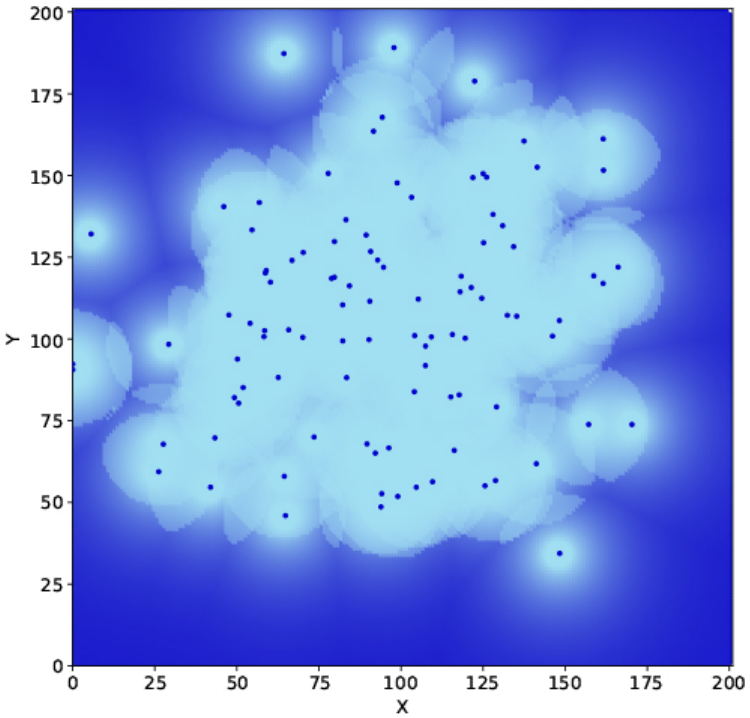}
		}\hspace{-0.005\linewidth}
		\subfigure[Gaussian deployment (optimal)]
		{
			\label{gaussian_opt}
			\includegraphics[width=0.22\linewidth]{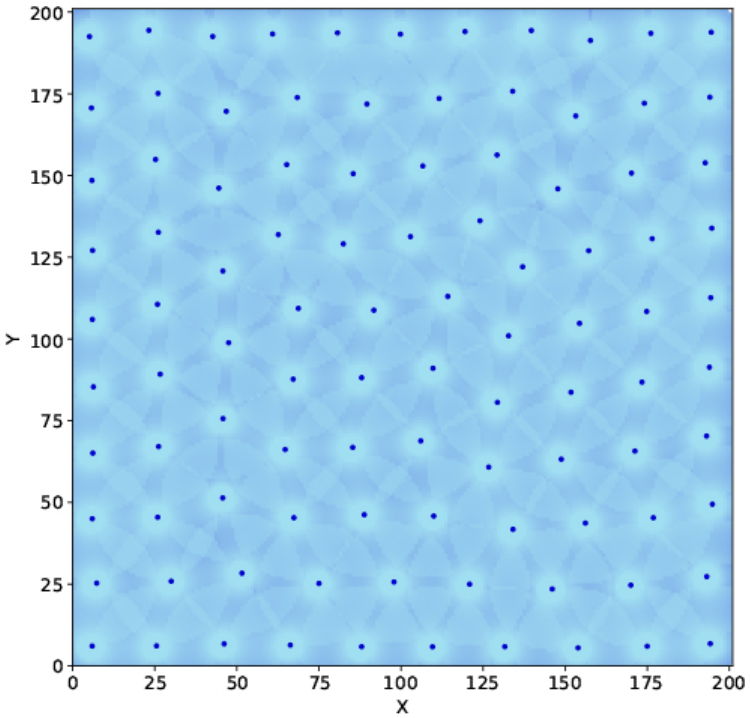}
		}\hspace{-0.005\linewidth}
		\subfigure[Logistic deployment (initial)]
		{
			\label{logistic}
			\includegraphics[width=0.22\linewidth]{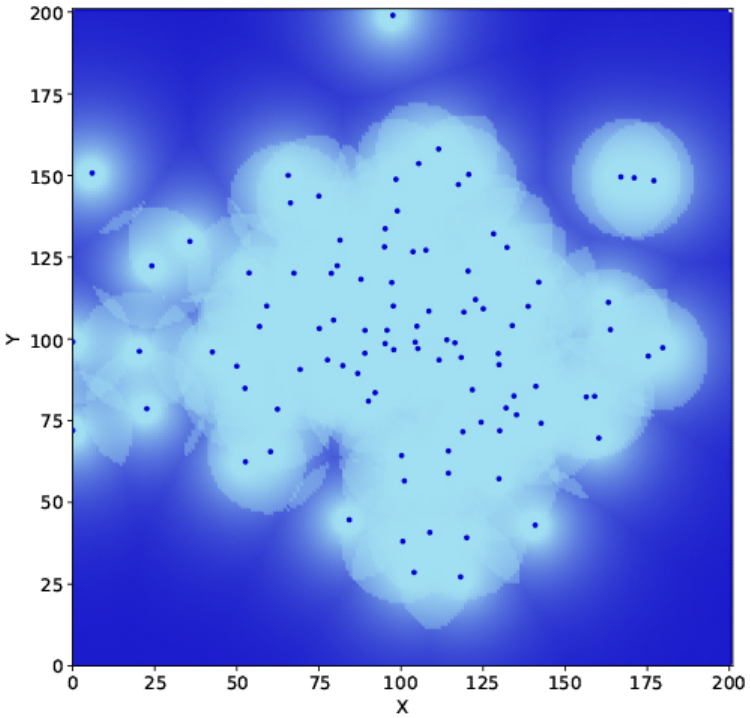}
		}\hspace{-0.005\linewidth}
		\subfigure[Logistic deployment (optimal)]
		{
			\label{logistic_opt}
			\includegraphics[width=0.22\linewidth]{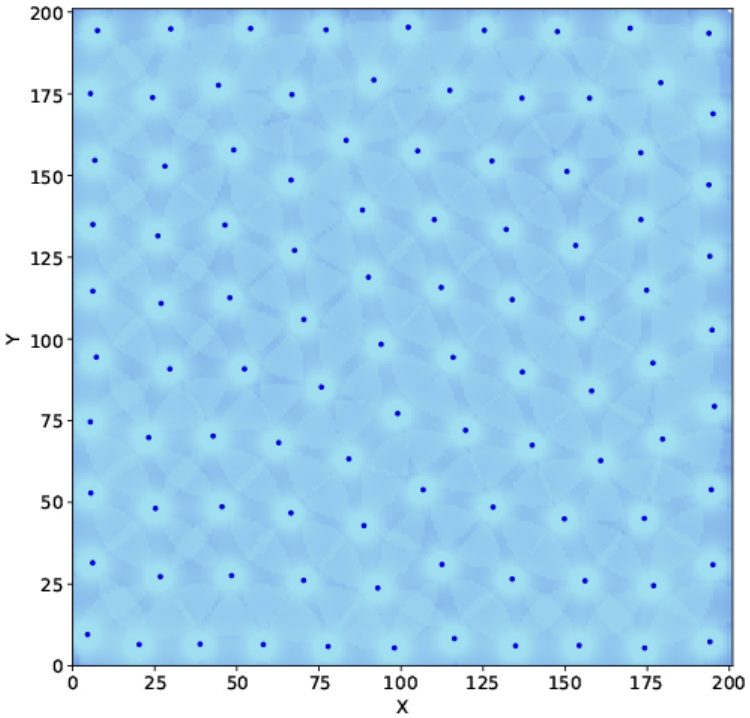}
		}
		\subfigure[Uniform deployment (initial)]
		{
			\label{uniform}
			\includegraphics[width=0.22\linewidth]{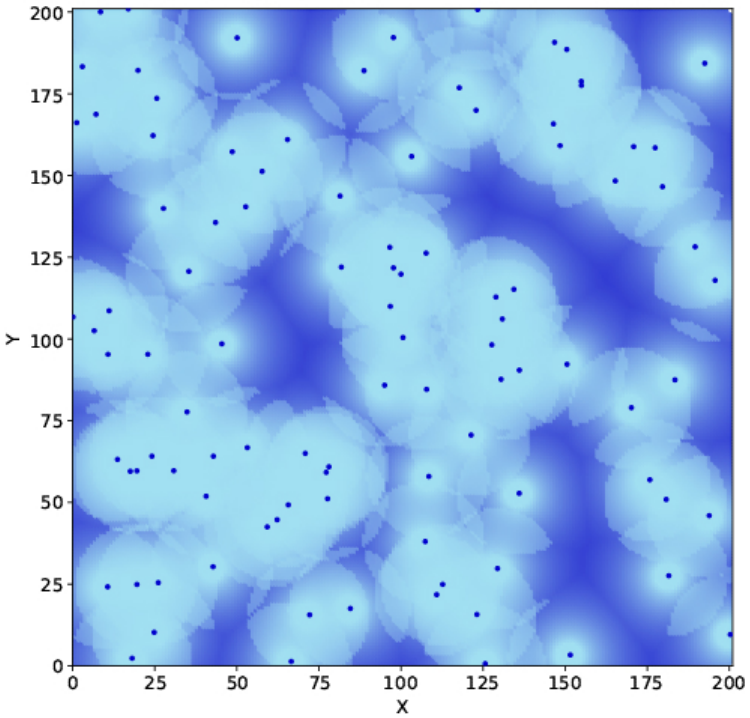}
		}\hspace{-0.005\linewidth}
		\subfigure[Uniform deployment (optimal)]
		{
			\label{uniform_opt}
			\includegraphics[width=0.22\linewidth]{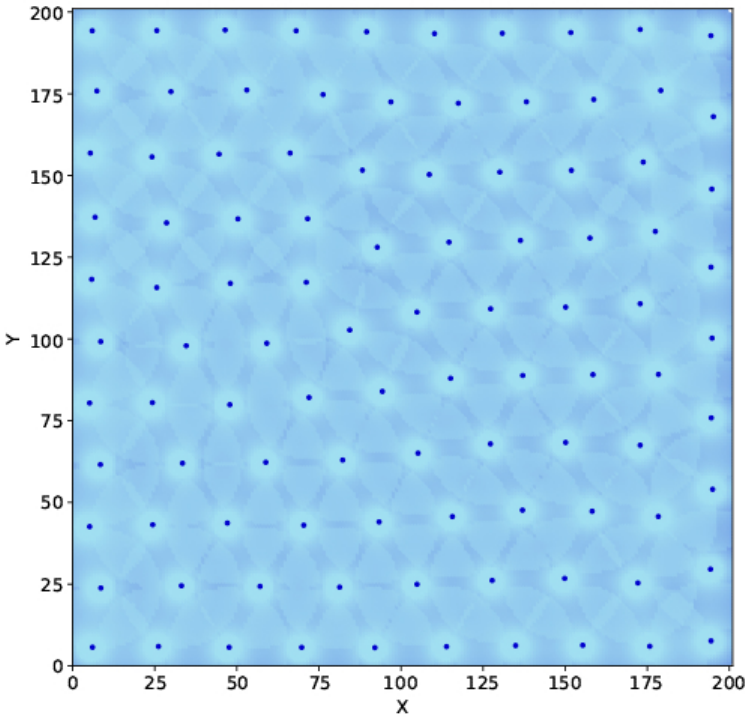}
		}\hspace{-0.005\linewidth}
		\subfigure[Exponential deployment (initial)]
		{
			\label{exponential}
			\includegraphics[width=0.22\linewidth]{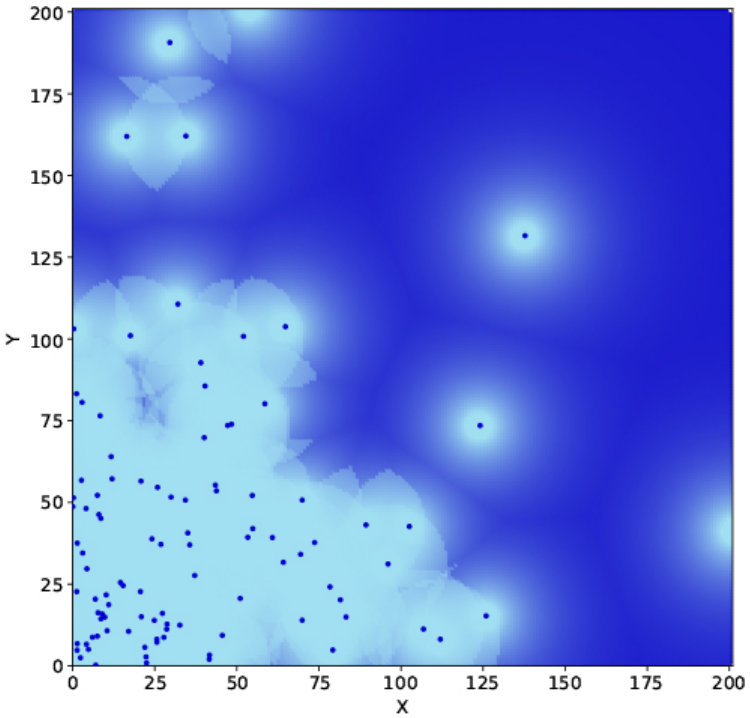}
		}\hspace{-0.005\linewidth}
		\subfigure[Exponential deployment (optinal)]
		{
			\label{exponential_opt}
			\includegraphics[width=0.22\linewidth]{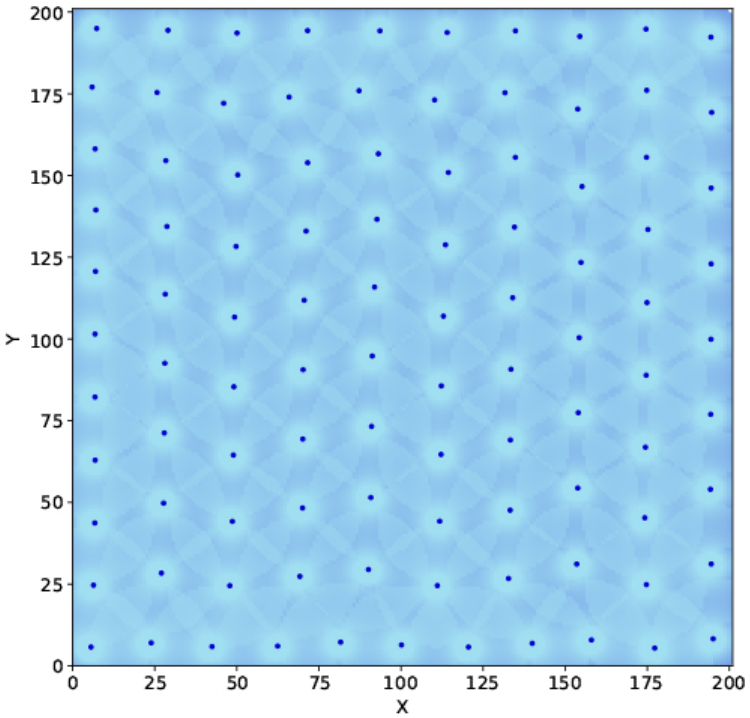}
		}

		\subfigure
		{
			\label{colorbar}
			\includegraphics[width=0.55\linewidth]{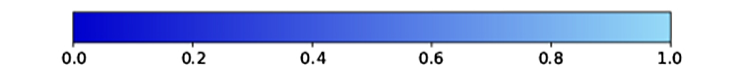}
		}\vspace{-0.01\linewidth}
		\caption{Optimization of WSN deployment using LSDNet-based algorithm with different initial deployment patterns, when $\mu=100$, $\sigma_G=35$, $\sigma_L=20$, $a=0$, $b=200$, and $\varLambda=\frac{1}{40}$.}
		\label{fig7}
	\end{figure*}

In Table \ref{tab3}, the centroid deployment refers to scattering all nodes around the centroid of the entire region, i.e., ($100,100$). The boundary deployment refers to scattering all nodes along the boundary of the entire region. The parameters of PDFs of other deployment patterns are set to $\mu=100$, $\sigma_G=35$, $\sigma_L=20$, $a=0$, $b=200$, and $\varLambda=\frac{1}{40}$. The detailed parameters of the LSDNet-based algorithm is identical as Table \ref{tab1}. Through optimization using the LSDNet, the initial and optimal deployment of 100 sensors are depicted in Fig. \ref{fig7}. The intensity of color of each grid represents the probability that the targets are collaboratively detected, as detailed in the color bar at the bottom of Fig. \ref{fig7}.

From Fig. \ref{fig7}, it can be observed that the optimization of WSN deployment using the LSDNet-based algorithm achieves the maximum coverage rate in all initial deployment patterns. Furthermore, the optimized sensors are relatively uniformly deployed within the region, ensuring the balanced contributions of sensors in detection. This experiment demonstrates that the  LSDNet-based algorithm exhibits robustness across diverse initial deployment patterns of sensors.

\subsubsection{Time Complexity Analysis}
LSDNet-based deployment outperforms other algorithms for its unique modeling. Assume the numbers of sensors and targets are $K$ and $N$, respectively. The time complexity of the filter layer is $O(K)$. For the evaluator layer, the distance calculation and the generalization of mass functions are both linear operations, so the time complexity is $O(K\cdot N)$. Additionally, Appendix \ref{appb} has demonstrated that the evidential fusion for any two mass functions of detection can be simplified to \eqref{eq36}, which leads to a decrease in time complexity from exponential to linear. Therefore, the time complexity of the evaluator layer and the fusion layer is $O(K\cdot N)$. In summary, the time complexity of the LSDNet-based deployment, i.e., Algorithm \ref{alg1}, approaches $O(K\cdot N)$. It is further verified by the linear increment in time consumption of LSDNet shown in Table \ref{tab2}.

For Algorithm \ref{alg2}, the time complexity of the calculation of the minimum number of sensors $K^{min}_{MBR}$ is $O(1)$. For the first $repeat$ loop, the number of redundant sensors decreases rapidly, so the number of loops $n_r$ is relatively small, i.e., $n_r \ll K$. For the second $repeat$ loop, the time complexity of the algorithms for finding the sensor with the most neighbors and removing the sensor and its coordinate elements is linear, i.e., $O(K)$. Therefore, the time complexity of the second $repeat$ loop is $O(K\cdot K^{min}_{MBR})$. In summary, the time complexity of the minimum sensors acquisition algorithm is $O(K\cdot (N+K^{min}_{MBR}))$.

\subsection{Irregular Forest Area Monitoring}\label{subsec5b}
\begin{figure}[h]
	\centering  
	\includegraphics[width=7cm]{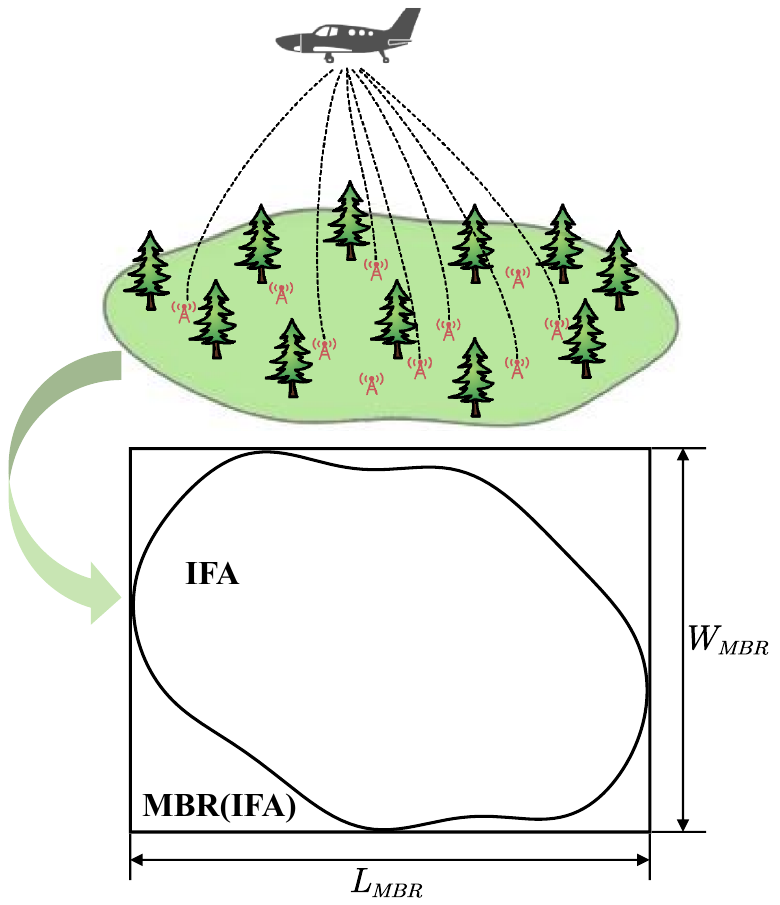}  
	\caption{An irregular forest area to be monitored.}  
 \label{fig8}
\end{figure}

In this section, we implement the LSDNet to
realize the WSN deployment of a real-world case,
i.e., the irregular forest area (IFA) monitoring, which is depicted in Fig. \ref{fig8}. The area of IFA is approximately $67840 \ \text{m}^2$. Due to the inaccessibility of the forest, a transport plane is employed to randomly scatter several wireless sensors with an identical sensing range $r_s=15\ \text{m}$, which comprise a WSN for monitoring purposes. Each sensor possesses self-locating abilities and is embedded with mobile devices. In order to achieve the maximum coverage of IFA as well as the minimum redundancy of sensor detection capabilities, wireless sensors are relocated to their optimal locations according to the LSDNet-based algorithm. For the convenience of analysis, the elevation difference of IFA is neglected, namely, the sensors are deployed in a two-dimensional space. 

\subsubsection{Minimum Sensors Acquisition for IFA Full Monitoring}
\begin{figure}[htpb]
	\centering  
	\includegraphics[width=6.5cm]{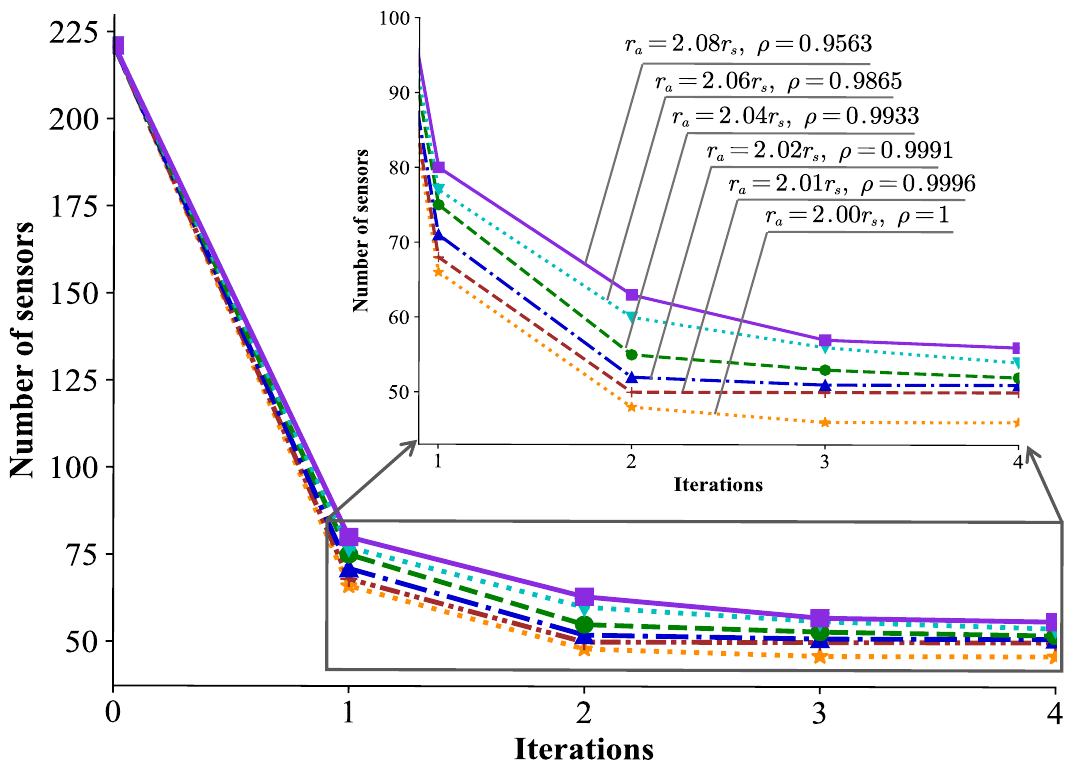}  
	\caption{Variation of sensors number and the maximum coverage rate when applying the minimum sensors acquisition algorithm.}  
 \label{fig9}
\end{figure}

\begin{figure}[htpb]
	\centering  
	\includegraphics[width=6.5cm]{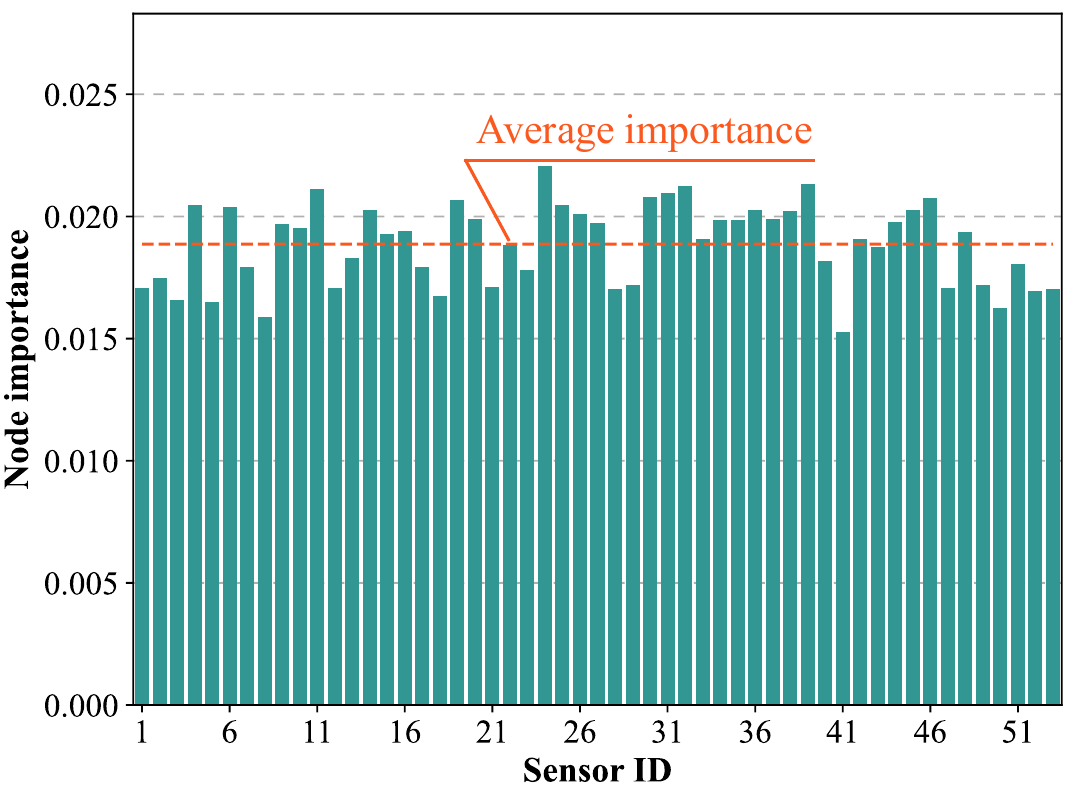}  
	\caption{The node importance of the optimal deployed sensors when $r_a=2.02r_s$.}  
 \label{fig10}
\end{figure}

Essentially, it is imperative to realize the full coverage of IFA for monitoring the entire region. Using the technique of distance measurement, the length and the width of the minimum bounding rectangle of IFA (indicated as $L_{MBR}$ and $W_{MBR}$ in Fig. \ref{fig8}), is $349 \ \text{m}$ and $261 \ \text{m}$, respectively.
Therefore, the minimum number of sensors required to achieve the full coverage of MBR(IFA) is computed as:
\begin{equation}\label{eq34}
\begin{split}
    K_{MBR}^{min}&=\lceil \frac{L_{MBR}}{\sqrt{2}r_s}\rceil \times \lceil\frac{W_{MBR}}{\sqrt{2}r_s}\rceil \\
    &=\lceil \frac{349}{\sqrt{2}\times 15}\rceil \times \lceil\frac{261}{\sqrt{2}\times 15}\rceil \\
    &=221
\end{split}
\end{equation}
which is an upper bound of the minimum number of sensors to fully cover the IFA. Algorithm \ref{alg2} is, then, employed to remove redundant sensors until the number of sensors reaches minimum that achieves the full coverage of IFA using the LSDNet. Thereby, by setting the overlapping radius $r_a\in[2r_s,2.08r_s]$, Fig. \ref{fig9} depicts the variation of number of sensors and the corresponding maximum coverage rate when applying the minimum sensors acquisition algorithm. When sensors exhibit greater repulsion to their neighbor sensors, i.e., the larger the overlapping radius $r_a $, the fewer sensors will be retained. Specifically, the full coverage of IFA can be achieved when the overlapping radius $r_a\leq 2.02r_s$ (within an error of $10^{-3}$). Therefore, the minimum number of sensors to achieve the full coverage of IFA using the LSDNet is 53. Accordingly, as illustrated in Fig. \ref{fig10}, the node importance of sensors of the optimal deployment when $r_a=2.02r_s$ falls around the level of average importance, indicating that the contribution of sensors in detecting targets within the IFA is properly balanced.

\subsubsection{Optimal WSN Deployment of IFA with Limited Sensors}
\begin{figure}[h]
		\centering  
		\subfigtopskip=0pt 
		\subfigbottomskip=0pt 
		\subfigcapskip=-5pt 
		\subfigure[]
		{
			\label{fig11a}
			\includegraphics[width=0.48\linewidth]{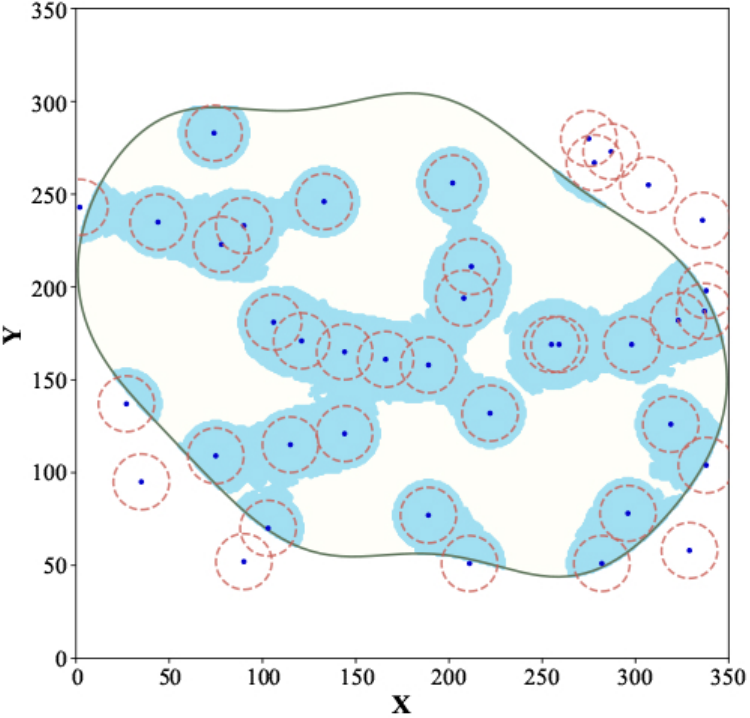}
		}\hspace{-0.04\linewidth} 
		\subfigure[]
		{
			\label{fig11b}
			\includegraphics[width=0.48\linewidth]{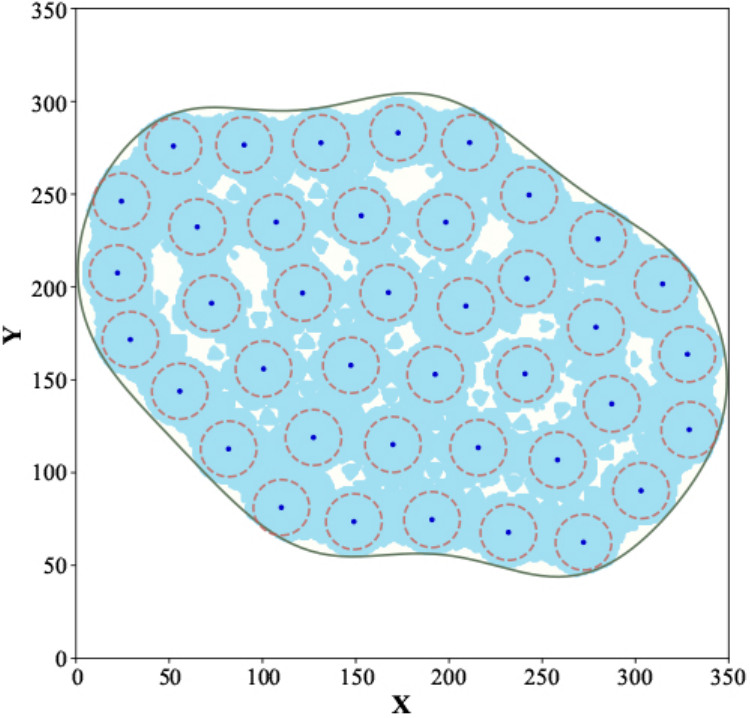}
		}\qquad
  		\subfigure
		{
			\label{cov_legend}
			\includegraphics[width=0.4\linewidth]{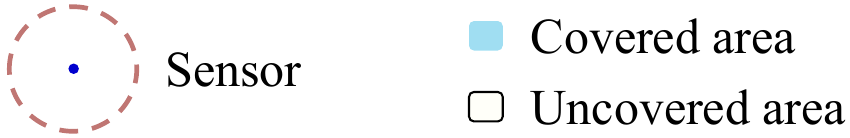}
		}
		\caption{Coverage range of WSN deployment using the LSDNet-based algorithm: (a)initial deployment; (b)optimal deployment.}
		\label{fig11}
	\end{figure}

\begin{figure*}[t]
	\centering
    \includegraphics[width=\textwidth]{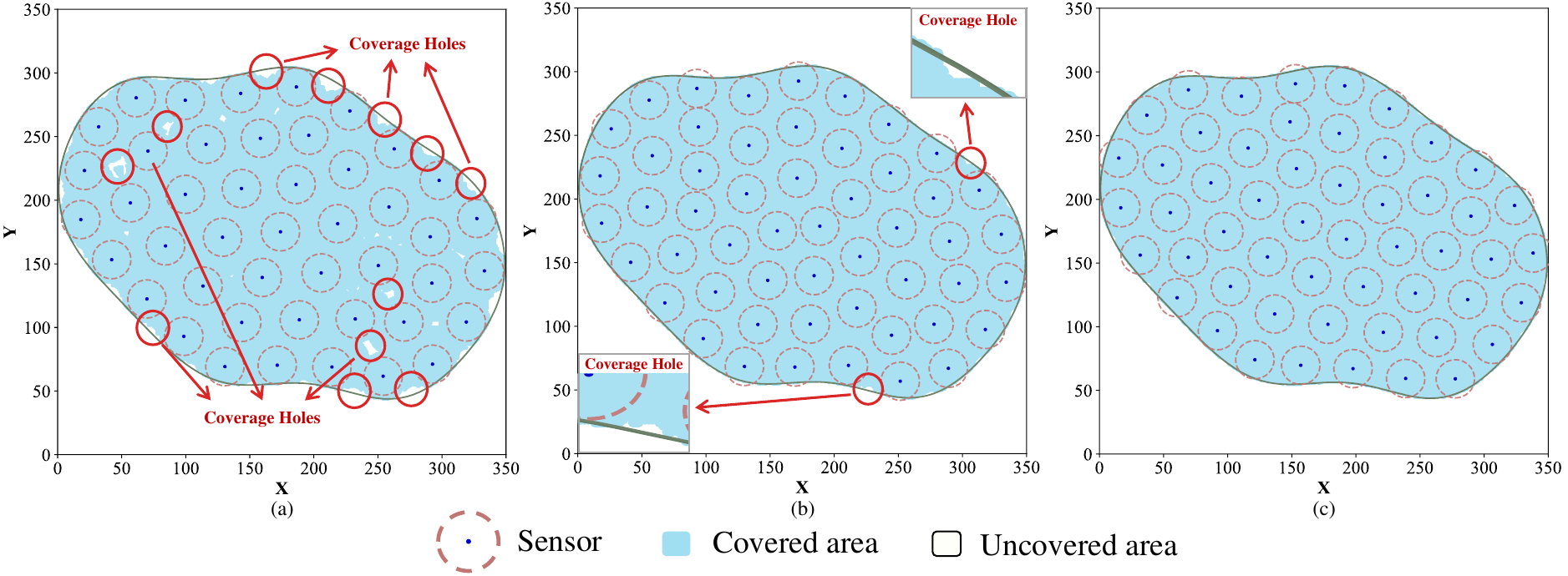}  
	\caption{Coverage areas using different sensor numbers: (a)$45$; (b)$52$; (c)$53$.}
	\label{fig12}
\end{figure*}

On most occasions, the quality of coverage of IFA cannot reach its maximum value when sensors are in short supply due to limitations of resources, e.g., budget, energy, and manpower. Moreover, the scatter of sensors tends to be stochastic, potentially resulting in the repeated coverage of one area, while other areas may be uncovered. As depicted in Fig. \ref{fig11a}, totally 40 sensors (cannot reach the maximum coverage quality) are randomly deployed around the IFA. The initial coverage quality is poor due to the irrational deployment, with some sensors even being scattered outside the IFA. In such circumstances, the LSDNet separates overlapping sensors and relocates those scattered outside the IFA back into the region to maximize the coverage rate, as illustrated in Fig. \ref{fig11b}. Specifically, the actual coverage area of the WSN exceeds the area directly covered by individual sensors, as the detection capability is determined by the probability of collaborative detection from multiple sensors rather than that of a single sensor. However, coverage holes remain inevitable when the number of sensors is insufficient. Fig. \ref{fig12} indicates the coverage areas under different sensor numbers. As the number of sensors increases, the size and number of coverage holes gradually diminish, eventually disappearing completely when the sensor number reaches 53.

\begin{figure}[h]
	\centering  
	\includegraphics[width=8cm]{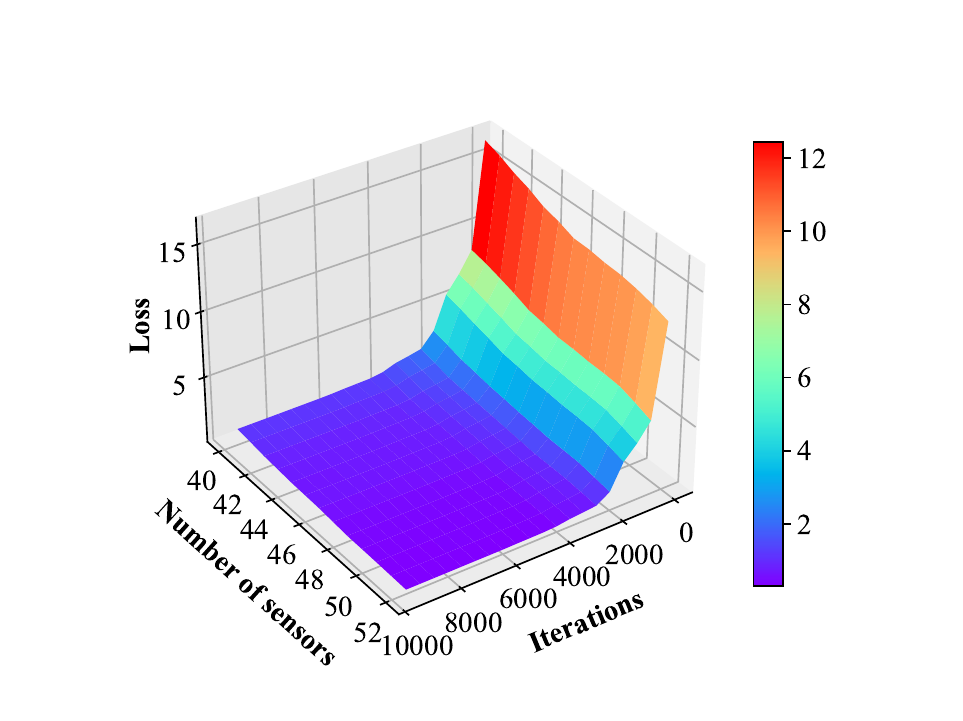}  
	\caption{The loss of the LSDNet-based optimization.}
 \label{fig13}
\end{figure}

Furthermore, by setting configurations with $N\in [40,52]$, Fig. \ref{fig13} illustrates the tendency of the loss of the LSDNet-based optimization. It shows that the loss is inversely proportional to the number of deployed sensors, which aligns with intuition. The solutions rapidly converge to the optima at the early stage of optimization, demonstrating the high performance of the proposed LSDNet-based optimization in terms of convergence.

\subsubsection{Selection of Initial Deployment Strategies}
\begin{figure}[h]
	\centering  
	\includegraphics[width=8.5cm]{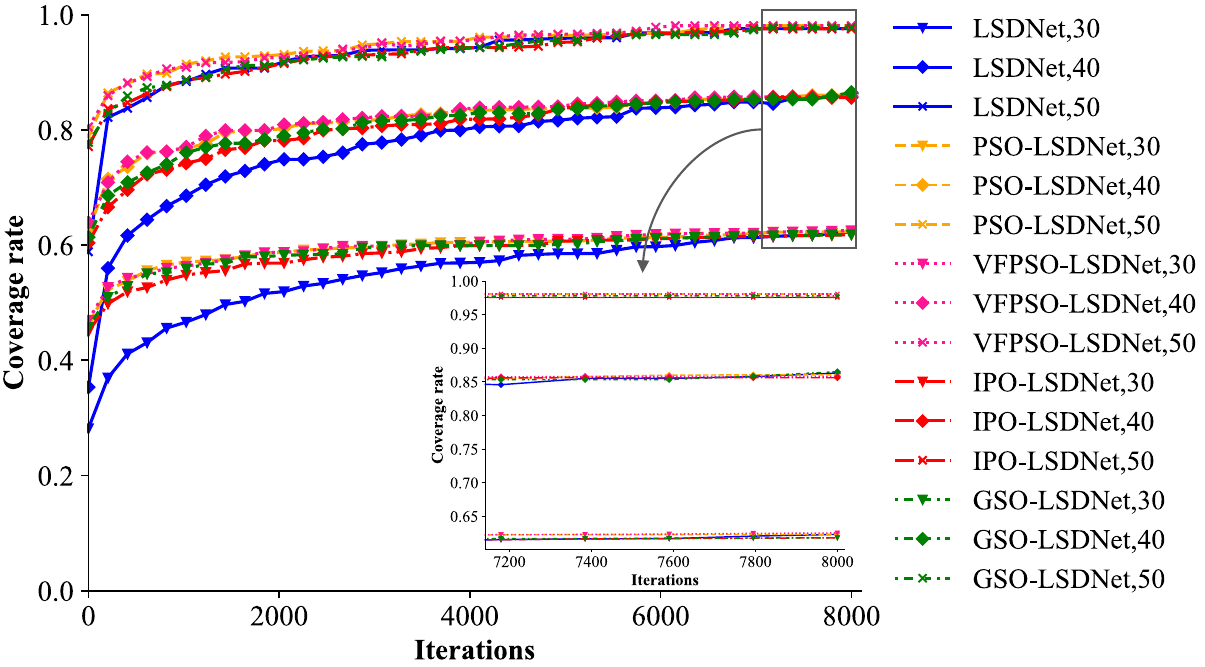}  
	\caption{The coverage rate of the IFA of LSDNet-based optimization, both with and without pre-deployment strategies.}
 \label{fig14}
\end{figure}

For an in-depth investigation, the relation between the initial deployment strategies and the deployment performance is illustrated in Table \ref{tab4} and Fig. \ref{fig14}. Specifically, Table \ref{tab4} depicts the mean and standard deviations of the number of iterations required by the minimum sensors acquisition algorithm with different deployment patterns via 20 independent executions. Obviously, random deployment, Gaussian deployment, Logistic deployment, and uniform deployment require fewer iterations to achieve convergence compared to other patterns. This suggests that greater spatial dispersion in WSN deployment leads to faster convergence of the algorithm. In particular, boundary deployment scatters most sensors outside the IFA, resulting in more iterations needed for convergence of the algorithm. Despite this, all of deployment patterns successfully obtain the minimum number of sensors, demonstrating the robustness of the proposed minimum sensors acquisition algorithm.
\begin{table*}[htpb]\small
\centering
\caption{Iterations of the minimum sensors acquisition algorithm with different deployment patterns.}
\label{tab4}
\tabcolsep=0.2cm
\renewcommand{\arraystretch}{1.1}
\begin{tabular}{cccccccc}
\toprule
Deployment patterns & Random & Centroid & Boundary & Gaussian & Logistic & Uniform & Exponential \\
\midrule
Number of iterations&4.3$\pm$0.3&5.4$\pm$0.4&6.2$\pm$0.5&4.4$\pm$0.3&4.4$\pm$0.2&4.1$\pm$0.1&4.8$\pm$0.4\\
Minimum sensors&53&53&53&53&53&53&53\\
\bottomrule
\end{tabular}
\end{table*}

The real-time performance is oftentimes considered a crucial metric for evaluating a WSN deployment algorithm. To eliminate the impact of sensor deployment patterns on deployment speed, we propose incorporating machine learning techniques into the initial sensors deployment process. Specifically, four pre-deployment strategies are implemented before using LSDNet to optimize WSN deployment. Fig. \ref{fig14} illustrates the increment in coverage rate achieved by implementing LSDNet alone and by implementing LSDNet following the pre-deployment phase, for deploying the WSN in the IFA. This case considers deploying 30, 40, and 50 sensors, with PSO, VFPSO, IPO, and GSO algorithms used during the pre-deployment phase, respectively. The results reveal that pre-deployments significantly increase the initial coverage rate of WSNs, which leads to nearly double the convergence speed of deployment algorithms compared to scenarios without pre-deployment. It is worth noting that the final coverage achieved is almost the same in all cases. It suggests that the use of machine learning techniques for pre-deployment reduces the dependency on initial sensors deployment and effectively improves the convergence speed of the LSDNet-based optimization of WSNs.

\section{Conclusion}
\label{sec6}
In this article, a learnable framework of WSN deployment was put forth to achieve the optimal quality of coverage via calibrating the sensors' locations. From the perspective of information fusion, detection capabilities of WSNs were enhanced by building collaborative sensing systems among multiple sensors using Dempster combination rule. The performance evaluation of such information fusion systems was tailored for selection of sensors for fusion. Then, we designed two LSDNet-based algorithms to address the optimization issues with respect to WSN deployment. To find the minimum number of sensors required for the full coverage of WSNs, we leveraged the greedy algorithm to remove redundant sensors. When possessing a limited number of sensors, the LSDNet was employed to equilibrate the node importance of sensors and maximize the coverage rate, which, thereby, yields optimal coverage quality of WSNs. The illustrative examples demonstrated the superiority of the proposed collaborative sensing model and the LSDNet in WSN deployment.

It is worth noting that the proposed framework and the related algorithms also have some limitations. First, some indicators (\textit{e.g.}, the energy consumption) are not considered in the LSDNet-based algorithm, which may result in the waste of resources. One of the promising solutions is to integrate the LSDNet with hierarchical clustering approaches to prolong the network lifetime and enhance the detection capabilities of WSNs. Second, the sensors in the proposed model own the identical parameters, which, nevertheless, may exhibits heterogeneity in real-world scenarios, e.g., the sensing range, the attenuation coefficient, and the lifetime. Considering this, rational integration of the detection information from heterogeneous sensors as well as their deployment strategies are worth exploring.

\appendices

\section{Proof of the Uncertainty monotonicity of the mass function of a single sensor}\label{appa}

\begin{proof}
According to the definition in Section \ref{subsec3b}, the distances between the target $t_j$ and the sensors satisfy $d(s_1,t_j)\leq d(s_2,t_j)\leq \cdots \leq d(s_k,t_j)$ $(k=1,2,\ldots,K)$. Therefore, the mass functions to detect the target $t_j$ satisfy $m^{\{D,ND\}}_{1,j}\leq m^{\{D,ND\}}_{2,j}\leq \ldots \leq m^{\{D,ND\}}_{k,j}$. Using \eqref{eq11} and \eqref{eq14}, the corresponding Hartley entropy of the mass function can be computed as:
\begin{equation}\label{eq35}
\begin{split}
	E_H(M_{k,j})=m^{\{D,ND\}}_{k,j}log(3)
\end{split}
\end{equation}
Therefore, Hartley entropy of the mass function satisfies $E_H(M_{1,j})\leq E_H(M_{2,j})\leq \cdots \leq E_H(M_{k,j})$, namely, the uncertainty of the mass function of detection of a single sensor is nondecreasing.
\end{proof}

\section{Proof of the Uncertainty monotonicity of the mass function of the collaborative sensing system}\label{appb}

\begin{proof}\label{proof2}
Assume that the mass function of the collaborative sensing system $k-1$ $(k\geq 3)$ is denoted as $M_{1\ldots(k-1),j}=(p,0,1-p)$ $(0< p \leq 1)$. The mass function of detection of the $k$th sensor is denoted as $M_{k,j}=(q,0,1-q)$ $(0< q \leq 1)$. Therefore, the mass function of the collaborative sensing system $k$ can be computed as:
\begin{equation}\label{eq36}
\begin{split}
	M_{1\ldots k,j}=(p-pq+q,0,1-p+pq-q)
\end{split}
\end{equation}
Using \eqref{eq11} and \eqref{eq14}, Hartley entropy of the mass function of the collaborative sensing system $k-1$ and $k$ are calculated by:
\begin{equation}\label{eq37}
\begin{split}
	E_H(M_{1\ldots (k-1),j})=(1-p)log_2(3)
\end{split}
\end{equation}
\begin{equation}\label{eq38}
\begin{split}
	E_H(M_{1\ldots k,j})&=(1-p+pq-q)log_2(3)\\
 &=[1-p+q(p-1)]log_2(3)
\end{split}
\end{equation}
It is obvious that $\forall 0< p \leq 1 \ and\ 0< q \leq 1,\ E_H(M_{1\ldots k,j})-E_H(M_{1\ldots (k-1),j}) \leq 0$. Therefore, $\forall k \geq 3,\ E_H(M_{1\ldots k,j})\leq E_H(M_{1\ldots (k-1),j})$, namely, the uncertainty of the mass function of the collaborative sensing system is nonincreasing  when fusing any sensor's mass function of detection.
\end{proof}

\section{Proof of the Boundness and the extremum condition of the efficiency}\label{appc}

\begin{proof}
When $0< p_e(s_k,t_j)<1$ $(k=2,3,\ldots,K)$, it can be readily proved that the term $\frac{E_{H}^{f}(M_{1\ldots k,j})}{E_{H}^{g}(M_{1,j},\ldots,M_{k,j})}> 0$, and the efficiency $\eta_H^{k,j}<1$. According to the denotations in Proof \ref{proof2}, the efficiency is denoted as:
\begin{equation}\label{eq39}
\begin{split}
\eta_H^{k,j} &= 1-\frac{E_{H}^{f}(M_{1\ldots k,j})}{E_{H}^{g}(M_{1,j},\ldots,M_{k,j})}\\
&= 1-\frac{E_{H}^{f}(M_{1\ldots k,j})}{\sqrt{E_H^f(M_{1\ldots (k-1),j})\times E_H(M_{k,j})}}\\
&= 1-\frac{(1-p-q+pq)log_2(3)}{\sqrt{(1-p)log_2(3)\times (1-q)log_2(3)}}\\
&= 1-\frac{(1-p)(1-q)}{\sqrt{(1-p)\times (1-q)}}\\
&= 1-\sqrt{(1-p)\times (1-q)}> 0
\end{split}
\end{equation}
It is noteworthy that $\eta_H^{k,j}=0$ holds if and only if $p=0$ and $q=0$. Actually, the infinite sensor-target distance cannot be reached, i.e., $p,q\neq0$. In such instance, thereby, $0<\eta_H^{k,j}<1$. When $p_e(s_1,t_j)=1$, specially, the target $t_j$ is completely detected by the first sensor, namely, the target $t_j$ is within the sensing range $r_s$, while the other sensor can be inefficient to detect it. Therefore, the efficiency of the collaborative sensing model is 0. Therefore, the proposed efficiency is bounded, i.e., $0 \leq \eta_H^{k,j}< 1$.
\end{proof}

\bibliographystyle{IEEEtran}  
\bibliography{references}

\end{document}